\documentclass[a4paper,11pt]{article}

\usepackage{makecell}

\usepackage{pgfplots}
\usepackage{graphicx}
\usepackage{lipsum}

\usepackage{algorithm}
\usepackage{algpseudocode}

\usepackage{geometry} 
\usepackage{indentfirst} 

\usepackage{setspace} 
\usepackage{amsmath}  
\usepackage{amsfonts} 
\usepackage{amssymb}  
\usepackage{bbm}

\newtheorem{definition}{Definition}[section]
\newtheorem{theorem}{Theorem}[section]
\newtheorem{proposition}{Proposition}[section]

\usepackage{titlesec}
\titleformat{\section}
{\normalfont\large}{\thesection{}.}{1.25em}{}

\titleformat{\subsection}
{\normalfont\large}{\thesubsection{}.}{0.5em}{}

\date{}

\title{Color Image Recovery Using Generalized Matrix Completion over Higher-Order Finite Dimensional Algebra}
\author{Liang Liao$^{1}$, Zhuang Guo$^{1}$, Qi Gao$^{1}$, Yan Wang$^{1}$, Fajun Yu$^{1}$, \\Qifeng Zhao$^{1}$, Stephen John Maybank$^{2}$}

\begin{document}

\maketitle
\begin{center}

{\small
$^{1}$~School of Electronics and Information, Zhongyuan University of Technology, Zhengzhou 451191, China\\
$^{2}$~Birkbeck College, University of Londn  WC1E 7HY, UK \\
}

\vspace{2em}
\small
liaoliang@ieee.org, sjmaybank@googlemail.com
\end{center}

\begin{abstract}
To improve the accuracy of color image completion with missing entries, we present a recovery method based on generalized higher-order scalars. We extend the traditional second-order matrix model to a more comprehensive higher-order matrix equivalent, called the ``t-matrix'' model, which incorporates a pixel neighborhood expansion strategy to characterize the local pixel constraints. This ``t-matrix'' model is then used to extend some commonly used matrix and tensor completion algorithms to their higher-order versions. We perform extensive experiments on various algorithms using simulated data and algorithms on simulated data and publicly available images and compare their performance. The results show that our generalized matrix completion model and the corresponding algorithm compare favorably with their lower-order tensor and conventional matrix counterparts.

\vspace{1em}
\end{abstract}

\section{Introduction}
\vspace{0.5em}
\subsection{Background and Related Works}
Vectors and matrices are fundamental to data analysis and processing, but they often struggle to encapsulate the complex, higher-order structures found in real-world applications such as color images, video sequences, and hyperspectral images. These multilinear data structures defy satisfactory representation by traditional vectors and matrices, prompting the use of tensors, higher-order extensions of vectors and matrices, for more accurate representation.

In real-world scenarios, it's common to find that higher-order, high-dimensional data often have low intrinsic dimensionality. This property facilitates several advanced techniques and applications. For example, Floryan et al. introduced a method to reduce data to their intrinsic dimensionality, allowing more accurate and low-dimensional dynamical models to capture the essential behavior of high-dimensional systems with low-dimensional features~\cite{floryan2022data}. Li et al. achieved efficiency and robustness by training deep neural networks in low-dimensional spaces without sacrificing performance~\cite{li2022low}. Chen et al. used deep learning networks for nonparametric regression on low-dimensional manifolds, emphasizing their adaptability to low-dimensional geometric structures in data~\cite{chen2022nonparametric}.

This notion of low intrinsic dimensionality often leads to the low rank or approximate low rank nature of higher order and high dimensional data when represented as matrices or tensors. Several methods take advantage of this property. Fu et al. developed a low-rank tensor approximation model for multiview intrinsic subspace clustering that effectively reduces view-specific constraints and improves optimization, with notable success on real-world datasets~\cite{Fu2022Low}. Wang et al.'s tensor low-rank and sparse representation method skillfully preserves intrinsic 3D structures in hyperspectral anomaly detection~\cite{wang2022learning}. In addition, Liu et al. comprehensively surveyed low-rank tensor approximation for hyperspectral image restoration, highlighting state-of-the-art techniques and current challenges in the field~\cite{liu2023survey}. This collective body of work underscores the flexibility and potential of using low-rank approximations to manage and interpret complex, high-dimensional data.

In addition, the collection of high-dimensional data can result in the loss of some elements. Low-Rank Tensor Completion (LRTC) addresses this problem by reconstructing the missing components from known data elements. Unlike Low-Rank Matrix Completion (LRMC), which relies solely on second-order information, LRTC exploits higher-order information, making the study of low-rank tensor completion techniques an important frontier in many fields.

For example, Liu et al. introduced the Sum of Matricization-based Nuclear Norms (SMNN), which is based on the Tucker rank of the tensor, and formulated three optimization algorithms for tensor completion via SMNN minimization, and successfully applied them to visual data completion~\cite{Liu2012Tensor}. Zhang et al. developed the Tubal Nuclear Norm (Tubal-NN) approach based on the Tubal-rank tensor, and designed an algorithm that uses tensor nuclear norm penalization for tensor completion, which proved to be effective in video recovery and denoising~\cite{Zhang2014Novel,kilmer2011factorization}. Lu et al. constructed a tensor completion model using the tensor nuclear norm (TNN), showed that TNN is a specific atomic norm, and established a bound for guaranteed low-tubal-rank tensor recovery, thus providing recovery guarantees for tensor completion~\cite{lu2018exact}. Xue et al. extended the tensor completion model to include the tensor truncated nuclear norm (T-TNN), thereby improving its effectiveness in real-world video and image processing~\cite{xue2018low}.

In addition to these newly defined tensor norms, researchers are now developing other techniques that extend traditional concepts to new applications in tensor completion. For example,Zeng introduced a multimodal nuclear tensor factorization technique, which incorporates low-rank insights and an efficient block successive upper bound minimization algorithm. The method was applied to tasks such as hyperspectral image, video, and MRI completion, with experimental results confirming its superior performance~\cite{zeng2022multimodal}. Similarly, Wu presented the tensor wheel decomposition method, a new tensor completion approach that characterizes complex interactions with fewer hyperparameters, improving performance on both synthetic and real data~\cite{wu2022tensor}. Zhao et al. presented a nonconvex model with a proximal majorization-minimization algorithm for robust low-rank tensor completion, providing theoretical guarantees and demonstrating high efficiency on visual data, including color and multispectral images~\cite{zhao2022robust}.

In recommendation systems, Deng et al. applied a meta-learning strategy with low-rank tensor completion for hyperparameter optimization and demonstrated its effectiveness~\cite{deng2022new}. 
Nguyen developed a consistency-based framework that emphasizes unit-scale consistency for matrix and tensor completion, with attributes such as fairness and the ability to exploit high-dimensional relationships~\cite{nguyen2023tensor}. Hui et al. integrated social-spatial context into tensor completion for time-aware point-of-interest recommendations, outperforming existing methods~\cite{hui2022time}.

Tensor completion has also contributed to advances in data mining. Song et al. reviewed recent tensor completion algorithms, examining four perspectives and various applications in data mining~\cite{song2019tensor}. Wu et al. introduced a multiattentional tensor completion network for handling missing entries in road sensor data, demonstrating improved performance~\cite{wu2022multi}. Lee's sign representable tensor model addressed both low and high rank signals for tensor completion, improving performance on human brain connectivity and topic data mining datasets~\cite{lee2021beyond}.

\subsection{Contributions and Organization of this Paper}

Building on the success of Low-Rank Tensor Completion (LRTC) over Low-Rank Matrix Completion (LRMC), this paper takes a leap forward by employing a higher-order t-matrix model with higher-order circular convolution. This novel method forms a specific generalization of LRMC tailored to multi-way image recovery challenges, including the completion of RGB images with missing entries.

Our model is inspired by and extends the well-regarded completion algorithm proposed by Lu et al. by incorporating a higher-order methodology that exploits the intricate relationships within high-dimensional data. Preliminary evaluations of our algorithm indicate that this generalized higher-order approach exhibits favorable recovery performance compared to existing algorithms.

Inspired by Lu et al.'s acclaimed completion algorithm, we present a higher-order methodology that exploits the inherent interrelationships within high-dimensional data. Evaluations of our approach demonstrate competitive recovery performance, with favorable recovery performance compared to existing algorithms.

The practical implications of this work are multifaceted, providing solutions that not only improve visual data completion, but also offer broader applications in areas such as video recovery, data mining, and medical imaging. By integrating our higher-order generalization into existing systems, it is possible to create more efficient and robust mechanisms for handling higher-order, high-dimensional data.

The contributions of this research can be summarized as follows. 
\begin{itemize}
	\item This research uses the higher-order t-matrix model, which generalizes low-rank matrix completion, to recover RGB images with missing entries. The model uses higher-order methodology to exploit complex relationships within high-dimensional data. The proposed method, termed ``Higher-order TNN'', compares favorably with its lower-order counterparts in terms of recovery performance, demonstrating distinct advantages.
	\item This research provides consistent solutions for visual data completion that have potential for broader applications. By integrating higher-order generalization into existing systems, it lays the groundwork for more effective analysis of higher-order, high-dimensional data.
	\item By generalizing the matrix model over finite-dimensional algebra, the approach employed extends several image analysis algorithms to their higher-order versions using a novel pixel neighborhood strategy.
	\item This research presents a consistent methodology for defining many of the notions of the t-matrix model, including rank, norm, and inner product, compared to the existing ones of the ``t-product'' model. This methodology provides insights into generalized scalars/matrices from the perspective of representation and operator theory. In addition, the study explores the application of higher-order Lagrange multipliers with generalized matrix variables.
\end{itemize}

The rest of the paper is organized as follows:
Section \ref{section:GeneralizedMatrices} introduces generalized matrices (t-matrices), outlining their structure, representation, and extension potential.
Section \ref{Low-RankMatrixCompletionandItsGeneralizations} describes the Low-Rank Matrix Completion (LRMC) methodology and its higher-order counterparts, covering mathematical formulations and generalizations.
Section \ref{section:RankConsiderations} provides an in-depth exploration of rank considerations, presenting different notions of rank and the concept of higher-order rank.
Section \ref{section:experiments} details experimental validation and performance analysis, using both simulated random data and real-world datasets such as the Berkeley segmentation dataset.
Section \ref{section:Conclusions} summarizes the content of this paper and its implications.
Appendix \ref{appendix:appendix} provides further mathematical justification, explaining the mechanism of t-scalars and t-matrices from a unique matrix perspective of representation and operator theory, along with an exploration of the Lagrange multiplier with t-matrix variables.

\section{Generalized Matrices}
\label{section:GeneralizedMatrices}
A generalized matrix (t-matrix) is a rectangular array composed of elements called generalized scalars (t-scalars)~\cite{liao2020generalized}. Since a generalized scalar forms an array in $\mathbbm{C}^{I_1 \times \cdots\times I_N}$, a generalized matrix with $D_1$ rows and $D_2$ columns can be represented by a multiway complex array in $\mathbbm{C}^{I_1 \times \cdots I_N \times D_1 \times D_2}$. While various authors, including Kilmer et al., categorize these generalized matrices as tensors~\cite{kilmer2011factorization,chang2022t,yu2023t}, we use the term ``generalized matrix over higher-order scalars''. Using this generalized matrix model provides an opportunity to extend many existing matrix algorithms.

\subsection{Generalized Scalars}
Let's consider a complex array of order $N$ to be an element of the set ${C}$, where ${C} \equiv \mathbbm{C}^{I_1 \times \cdots \times I_N}$. In parallel, a real array of order $N$ is identified as an element of the set ${R}$, where ${R} \equiv \mathbbm{R}^{I_1 \times \cdots \times I_N}$. The sets ${R}$ and ${C}$ share the commutative ring structure, where the multiplication of their elements is defined by the circular convolution of order $N$, and the addition corresponds to the entry-wise array addition. Elements inside ${C}$ and ${R}$ are called generalized scalars. In this paper, we focus primarily on ${C}$, since ${R}$ is a subset of ${C}$. By further defining the multiplication of a generalized scalar with a complex number by conventional scalar multiplication, we can elevate the ring ${C}$ to a finite dimensional commutative algebra.

Using generalized scalars not only allows us to construct novel matrices, but also extends many classical matrix algorithms into the realm of generalized matrix algorithms. In 2011, Kilmer et al. pioneered the ``t-product'' model~\cite{kilmer2011factorization}. In this model, scalar elements in traditional matrices are replaced by fixed-size one-dimensional arrays, allowing the extension of many classical matrix algorithms. Taking advantage of this extension, new generalized matrices use elements from the commutative algebras $R$ or $C$ - the generalized scalars - and are thus considered matrices built on the foundation of finite-dimensional commutative algebras.

In this paper, we adopt generalized matrices inspired by the work of Kilmer et al.~\cite{kilmer2011factorization}, following the model of Liao and Maybank~\cite{liao2020generalized}. This research, based on the notions introduced in the generalized matrix model~\cite{liao2020generalized}, called ``t-matrix'', extends Kilmer et al.'s order-one generalized scalars to higher orders through a neighborhood strategy, thereby extending conventional matrices to their high-order versions. Then, the multi-way circular convolution of generalized scalars in the spatial domain is translated into Hadamard multiplication in the Fourier domain via the Fourier transform, facilitating relevant computations. For example, the following definitions are given for generalized scalars.
\begin{definition}[Addition of Generalized Scalars~\cite{liao2020generalized}]
	\label{definition:67674543434}
	Consider two generalized scalars, called t-scalars, 
	$\dot{x}, \dot{y} \in \mathbbm{C}^{I_1 \times \dots \times I_N}$ as order-two arrays of size $I_1 \times \dots \times I_N$. The sum, $\dot{c} = \dot{x} + \dot{y}$, with $\dot{c} \in \mathbbm{C}^{I_1 \times \dots \times I_N}$, is calculated element-wise, meaning the complex entry of $\dot{c}$ at position $(i_1,\dots,i_N)$ is 
	$$
	(\dot{c})_{i_1,\dots,i_N} = (\dot{x})_{i_1,\dots,i_N} + (\dot{y})_{i_1,\dots,i_N} \quad\forall i_1,\dots,i_N.
	$$
\end{definition}

\begin{definition}[Multiplication of Generalized Scalars~\cite{liao2020generalized}]
	\label{definition:7867564534}
	Let $\dot{x}, \dot{y} \in \mathbbm{C}^{I_1 \times \dots\times I_N}$ be two t-scalars of size $I_1 \times \dots\times I_N$. Their product is defined as $\dot{c} = \dot{x} \circ \dot{y}$, where $\dot{c}$ results from the order-two circular convolution of $\dot{x}$ and $\dot{y}$. Specifically, we have
	\begin{equation}
		(\dot{c})_{i_1,\dots,i_N}=\sum_{(k_1,\dots,k_N) \in [I_1]\times \dots \times [I_N]}
		(\dot{x})_{k_1,\dots, k_N} \cdot
		(\dot{y})_{k_1^{\prime},\dots,k_N^{\prime}}, \quad\forall i_1,\dots,i_N
		\nonumber
	\end{equation}
	where $k_n^{\prime}=\operatorname{mod}(i_n - k_n, I_n)+1 $, for all $n \in [N]$  .
\end{definition}

While order-two generalized t-scalars share the data structure of an order-two numerical array, they are not matrices, since their multiplication is by definition commutative. However, to describe linear transformations, it can be convenient to consider the underlying order-two arrays as matrices. 

We use the notation $\operatorname{tensor}(\dot{x})$ to elevate the underlying order-$N$ array of $\dot{x}$ to a conventional tensor of identical size and entries.
With $\operatorname{tensor}(\dot{x})$ as the conventional tensor, the multiplication of two t-scalars is equivalently given by the following theorem.
\begin{theorem}[Fourier Transform]
	Let $\dot{x}, \dot{y} \in C$ be two t-scalars with the product $\dot{c} = \dot{x} \circ \dot{y} \in C$. Define $\tilde{x} = F(\dot{x})$, $\tilde{y} = F(\dot{y})$, and $\tilde{c} \doteq F(\dot{c})$ as their respective multilinear Fourier transforms. For any t-scalar $\dot{x} \in C$, its multilinear Fourier transform is given by the following multi-mode multiplication:
	\begin{equation}
		F(\dot{x}) \doteq
		\operatorname{tensor}(\dot{x}) \times_1 W_{1}  \dots  \times_{n} W_{n}  \dots  \times_N W_{N} 
		\label{equation:89675656}
	\end{equation}
	where $W_{n} $ denotes the Fourier matrices of appropriate size. Consequently, the following Hadamard product holds for all $i_1,\dots,i_N$:
	\begin{equation}
		(\tilde{c})_{i_1,\dots,i_N} = (\tilde{x})_{i_1,\dots,i_N} \cdot (\tilde{y})_{i_1,\dots,i_N} \;.
		\nonumber
	\end{equation}
\end{theorem}

Definitions \ref{definition:67674543434} and \ref{definition:7867564534} qualify all t-scalars as elements of a commutative ring $C$. 
An essential operation to bring the ring $C$ to life as an algebra is the multiplication of any element of $C$ with a conventional scalar. This leads to the following definition.
\begin{definition}[Scalar Multiplication~\cite{liao2020generalized}]
	Let $\dot{x} \in C \equiv \mathbbm{C}^{I_1 \times \dots\times I_N}$ be a generalized scalar and $\lambda$ a complex number. Their multiplication, denoted as $\dot{y} \doteq \lambda \cdot \dot{x} \in C$, is defined for all $i_1,\dots,i_N$ as follows:
	$$
	(\dot{y} )_{i_1,\dots,i_N}= \lambda \cdot  (\dot{x})_{i_1,\dots,i_N}   \;.
	$$
\end{definition}

With the previous definitions as a basis, we can easily have the identity and zero t-scalars within the algebra $C$. These two unique t-scalars are defined as
\begin{proposition}[Identity T-scalar and Zero T-scalar~\cite{liao2020generalized}]
	Consider a t-scalar $\dot{e} \in \mathbbm{C}^{I_1 \times \dots \times I_N}$. In this case, $(\dot{e})_{i_1,\dots,i_N}=1$ if $i_1 = \dots = i_N =1$, and $(\dot{e})_{i_1,\dots,i_N} = 0$ otherwise. The first entry is $1$, while all other entries are $0$, characterizing the identity t-scalar $\dot{e}$. Alternatively, if every entry of $\dot{x}$ is $0$, we have this t-scalar as the zero t-scalar $\dot{z}$. Note that each entry of the identity t-scalar is $1$ in the Fourier domain, while the zero t-scalar remains unchanged in the Fourier domain.
\end{proposition}

\subsection{T-scalars as Finite-dimensional Linear Operators}
\label{section:GeneralizedScalars}
Since every generalized scalar in the algebra $C$ operates as a finite-dimensional linear commutative operator, operator theory allows us to determine the spectrum of any t-scalar $\dot{x} \in C$. The spectrum, or the set of complex eigenvalues of $\dot{x}$, corresponds to the $K$ entries of the Fourier transform $\tilde{x}$ (where $K \doteq I_1\cdot I_2\cdots I_N$), taking multiplicity into account.

With the eigenvalues (i.e., Fourier entries) of an arbitrary t-scalar at hand, let us now delve into the following definitions.
\begin{definition}[Conjugate~\cite{liao2020generalized}]
	A unique t-scalar $\dot{y}$ in $C$ is the conjugate of an t-scalar $\dot{x}$ in $C$ if each eigenvalue of $\dot{y}$ is the complex conjugate of the corresponding eigenvalue of $\dot{x}$.  The conjugate is denoted by $\dot{x}^{*}$.
\end{definition}

\begin{definition}[Nonnegativity~\cite{liao2020generalized}]
	\label{definition:89785656}
	A t-scalar $\dot{x}$ is said to be nonnegative if and only if all of its complex eigenvalues (i.e., Fourier entries) are nonnegative real numbers.
\end{definition}

Definition \ref{definition:89785656} is crucial because it facilitates the generalization of various concepts of nonnegativity, including matrix rank, space dimension, norm, and distance, to nonnegative elements in $C$. We will explore these generalizations as needed.

The nonnegativity established in Definition \ref{definition:89785656} also establishes other important concepts. For example, consider any two nonnegative generalized scalars $\dot{x}, \dot{y}$. Their subtraction is called self-conjugate, since the equation $(\dot{x} - \dot{y})^{*} = \dot{x} - \dot{y}$ holds invariably. Furthermore, the element $\dot{z} \in C$ represents the smallest nonnegative element. If $\dot{x} - \dot{y}$ is nonnegative, a partial order $\dot{x} \geq \dot{y} \geq \dot{z}$ is defined. If neither $\dot{x} - \dot{y}$ nor $\dot{y} - \dot{x}$ is nonnegative, $\dot{x}$ and $\dot{y}$ are said to be incomparable.

\subsection{Generalized Matrices}
A generalized matrix, called a t-matrix, is a rectangular array of generalized scalars. Since these generalized scalars (i.e., t-matrices) are arrays in $\mathbbm{C}^{I_1 
	\times \dots\times I_N}$, it is logical to represent the underlying data form of a generalized matrix in $C^{D_1 \times D_2}$ as an order-four array in $\mathbbm{C}^{I_1 \times \dots \times I_N \times D_1 \times D_2}$. We refer to this format as the little-endian representation of a t-matrix. Conversely, some authors may arrange a t-matrix in $C^{D_1\times D_2}$ as an array in $\mathbbm{C}^{D_1\times D_2\times I_1\times \dots\times I_N}$ rather than in $\mathbbm{C}^{I_1\times \dots \times I_N\times D_1\times D_2}$. 
We call this form the big-endian representation, which is used by Kilmer et al. in their paper~\cite{kilmer2011factorization}. 
Despite the two protocols, the conversion between the little-endian and big-endian protocols is straightforward. Because of the underlying multiway array structure of t-matrices, some authors refer to these t-matrices as tensors~\cite{kilmer2013third}, although they are different from ordinary tensors with complex entries.

The operations on t-matrices are analogous to those on traditional matrices. Specifically, if we have a t-matrix in $C^{D_1\times D_2} \equiv \mathbbm{C}^{I_1\times \dots \times I_N\times D_1 \times	D_2}$ and another in $C^{D_2\times D_3} \equiv \mathbbm{C}^{I_1\times \dots \times I_N\times D_2\times D_3}$, their multiplication yields a t-matrix in $C^{D_1\times D_3} \equiv \mathbbm{C}^{I_1\times \dots \times I_N\times D_1 \times D_3}$.
Similarly, constructs such as the conjugate transpose and the diagonal matrix can be defined analogously. For a more detailed discussion of these concepts, see~\cite{liao2020generalized}.

\subsection{Singular Value Decomposition of a Generalized Matrix}
To exploit the structure of a generalized matrix, it is often decomposed into a sequence of simpler components, which is usually written in a matrix form. Parallel to the compact Singular Value Decomposition (SVD) of traditional matrices, a crucial decomposition, called TSVD (Tensorial SVD), of a generalized matrix $\dot{X} \in C^{D_1\times D_2}$ is shown below:
\begin{equation}
	\dot{X} = \dot{U} \circ \dot{S} \circ \dot{V}^{*}
	\label{equation:89786767}
\end{equation}
where $\dot{U} \in C^{D_1\times D}$, $\dot{S} \in C^{D \times D }$, and $\dot{V} \in C^{D_2 \times D }$, where $D \doteq \min(D_1, D_2)$. The symbol $\dot{V}^{*}$ denotes the conjugate transpose of the t-matrix $\dot{V}$, and $\dot{S} \doteq \operatorname{diag} 
(
\dot{\sigma}_1, \cdots,
\dot{\sigma}_{D}
)
$ is a diagonal t-matrix with nonnegative t-scalars as its diagonal entries. 
The partial order is
$\dot{\sigma}_1   \geq  \cdots \geq  \dot{\sigma}_D \geq \dot{z}$.

In addition, the following generalized orthogonal constraints are available:
\begin{equation}
	\dot{U}^{*} \circ \dot{U} = 
	\dot{V}^{*} \circ \dot{V} = \dot{I} \doteq \operatorname{diag} (
	\dot{e}, \dots,
	\dot{e})  \in C^{D\times D} \equiv \mathbbm{C}^{I_1\times \dots \times I_N\times D\times D}
\end{equation}
Here $\dot{I}$ denotes the identity t-matrix, which has diagonal entries of $\dot{e}$ and off-diagonal entries of $\dot{z}$.

Equation \ref{equation:89786767} defines the tensorial singular value decomposition (TSVD) of a t-matrix. Although its non-uniqueness persists, numerous methods expound on the computational and operational aspects of TSVD. Among these, one particularly practical method employs the mechanism of spectral slices.

Given a t-matrix $\dot{X} \in C^{D_1\times D_2}$, represented as a little-endian complex array in $\mathbbm{C}^{I_1\times \dots \times I_N\times D_1\times D_2}$, let $\operatorname{tensor}(\dot{X})$ map this underlying array into a conventional tensor with identical size and entries. Complying with Equation \ref{equation:89675656}, the Fourier transform of $\dot{X}$ can be expressed as the following multi-mode multiplication:
\begin{equation}
	\tilde{X} \doteq	F(\dot{X}) = 
	\operatorname{tensor}(\dot{X}) \times_1 W_{1}  \dots  \times_{n} W_{n}  \dots  \times_N W_{N} 
	\label{equation:89785656}
\end{equation}

Since all operations on t-scalars in the Fourier domain are Fourier-entry-wise, the following definition can be used to decompose t-matrices in the Fourier domain and to establish further constructs.

\begin{definition}[Spectral Slice~\cite{liao2020generalized}]
	For any t-matrix $\dot{X} \in C^{D_1\times D_2} \equiv \mathbbm{C}^{I_1\times \dots \times I_N\times D_1\times D_2}$ and its Fourier transform $\tilde{X} \in \mathbbm{C}^{I_1\times \dots\times I_N\times D_1\times D_2}$ as defined in Equation \ref{equation:89785656}, can be partitioned into $K$ spectral slices (where $K = I_1 \cdot I_2\cdots I_N$). Each spectral slice, indexed by $(i_1,\dots,i_N)$, is a conventional complex matrix denoted by $\tilde{X}(i_1,\dots,i_N)\in \mathbbm{C}^{D_1\times D_2}$. This satisfies the following equation for all $i_1,\dots,i_N$ and $d_1, d_2$:
	\begin{equation}
		\big(\tilde{X}(i_1,\dots,i_N) \big)_{d_1, d_2} = (\tilde{X})_{i_1,\dots,i_N, d_1, d_2} \;.
	\end{equation}
\end{definition}

Using spectral slices, various constructs can be introduced. For example, the Tensor Singular Value Decomposition (TSVD) of a generalized matrix is outlined in Algorithm \ref{alg:TensorSVD}. 
\begin{algorithm}
	\caption{Tensorial Singular Value Decomposition via Spectral Slices}
	\begin{algorithmic}[1]
		\Procedure{TSVD}{$\dot{X}$}~\vspace{0.2em}
		\State Apply Equation \ref{equation:89785656} to compute the transform $\tilde{X}$ from $\dot{X}$.~\vspace{0.2em}
		\For {$(i_1,\dots,i_N)  \in [I_1] \times \dots\times [I_N]$}~\vspace{0.2em}
		\State Compute the compact SVD of each spectral slice $\tilde{X}(i_1,\dots,i_N) = U \cdot S \cdot V^{H}$.~\vspace{0.2em}
		\State Store the resulting $U$, $S$, and $V$ in $\tilde{U}(i_1,\dots,i_N)$, $\tilde{S}(i_1,\dots,i_N)$, and $\tilde{V}(i_1,\dots,i_N)$.~\vspace{0.2em}
		\EndFor~\vspace{0.2em}
		\State Apply the inverse transform $F^{-1}$ to each of $\tilde{U}$, $\tilde{S}$, and $\tilde{V}$ to obtain $\dot{U}$, $\dot{S}$, and $\dot{V}$, 
		\Statex \hspace{1.5em}respectively.~\vspace{0.2em}
		\EndProcedure
	\end{algorithmic}
	\label{alg:TensorSVD}
\end{algorithm}

Spectral slices facilitate the extension of many conventional algorithms, including the acclaimed Singular Value Thresholding (SVT) algorithm, reported in ~\cite{Candes2011robust,candes2012exact}, and integral to Low Rank Matrix Completion (LRMC) problems. The special implementation of Singular Value Thresholding on generalized scalars is presented in Algorithm \ref{alg:TensorSVT}.
\begin{algorithm}
	\caption{Tensorial Singular Value Thresholding via Spectral Slices}
	\begin{algorithmic}[1]
		\Procedure{$\dot{Y}$ = TSVT}{$\dot{X}, \tau$} where $\tau$ is a small positive constant~\vspace{0.2em}
		\State Use Equation \ref{equation:89785656} to compute the transform $\tilde{X}$ of $\dot{X}$.~\vspace{0.2em}
		\For {$(i_1,\dots,i_N) \in [I_1] \times \dots\times [I_N]$}~\vspace{0.2em}
		\State Compute the compact SVD on each spectral slice $\tilde{X}(i_1,\dots,i_2) = U \cdot S \cdot V^{H}$.~\vspace{0.2em}
		\State Store the result of the singular value thresholding 
		$U \cdot C_{\tau}(S) \cdot V^{H}$ in $\tilde{Y}(i_1,\dots,i_N)$.~\vspace{0.2em}
		\EndFor~\vspace{0.2em}
		\State Apply the inverse transform $F^{-1}$ to $\tilde{Y}$ to obtain $\dot{Y}$, the approximation of $\dot{X}$.~\vspace{0.2em}
		\EndProcedure
	\end{algorithmic}
	\label{alg:TensorSVT}
\end{algorithm}


\section{Low-Rank Matrix Completion and Its Generalizations}
\label{Low-RankMatrixCompletionandItsGeneralizations}
Besides the generalization of SVD to its higher-order counterpart TSVD, many other matrix algorithms can be extended analogously over t-scalars. One of them is the so-called low-rank matrix completion (LRMC) problem.

\subsection{Low Rank Matrix Completion}
A variant of the matrix completion problem is to determine the minimum rank matrix $X \in \mathbbm{R}^{D_1\times D_2}$ that matches the desired matrix $M$ for all observed entries within the index set $\Omega$~\cite{candes2012exact}. This problem can be expressed mathematically as
\begin{equation}
	\mathop{\operatorname{minimize}}\limits_X \operatorname{rank}(X)
	\text { subject to } (X)_{i,j}= (M)_{i,j}\quad \forall (i, j) \in \Omega .
	\label{equation:787867567}
\end{equation}

Given the NP-hard nature of the initial minimization problem, in most practical scenarios the solution to Equation \ref{equation:787867567} can most likely be reformulated as the solution to the following convex optimization problem:
\begin{equation}
	\mathop{\operatorname{minimize}}\limits_{X, \,E}\|X\|_{*} \text{ subject to } X + E = M, ~G_{\Omega}(E) = \text{array of zeros} 
	\label{equation:convex}
\end{equation}
where $G_{\Omega}: \mathbbm{R}^{D_1\times D_2} \rightarrow \mathbbm{R}^{D_1\times D_2}$ represents the linear operator, which preserves entries within the set $\Omega$ and sets entries outside of $\Omega$ to zero.

The augmented Lagrange multiplier function for the minimization problem is formalized as
\begin{equation}
	L(X, E, Y, \tau)= \|X\|_{*}+ \langle Y, M-X-E \rangle + \frac{1}{2\tau} \|M - X- E \|_F^2
	\label{equation:lagrange}
\end{equation}
where $Y$ is the dual variable and $\tau > 0$.

The Alternating Direction Method of Multipliers (ADMM)~\cite{lin2022alternating,han2022survey} can be used to iteratively refine the optimization variables $X$ and $E$, as described in Algorithm \ref{alg:ADMM}. 
Here, $M \in \mathbbm{R}^{D_1\times D_2}$, and $\Omega$ represents a random non-empty proper subset of the Cartesian product $[D_1]\times [D_2]$. Furthermore, $D_{\tau}$ in line \ref{TradiionalADMM-step:4} denotes the SVT operator with threshold $\tau$. 
\begin{algorithm}
	\caption{ADMM for solving Equation \ref{equation:convex}}
	\begin{algorithmic}[1]
		\Procedure{$X_\mathit{COMP}$ = MatrixCompletion}{$M, \Omega$}  
		~\vspace{0.2em}
		\State Initialization: $k \gets 0$,  $Y_0 = E_0 \gets ~\text{array of zeros}$,  
		$\alpha \gets 0.9$, 
		$\tau_0 \gets 10^{4}$,  
		$\tau_\mathrm{min} \gets 10^{-6}$  
		\State Set the missing entries of $M$, i.e., $(i, j) \in \Omega^{c}$, to zero 
		~\vspace{0.2em}
		\While {neither convergence nor predefinite maximum iterations achieved}~\vspace{0.2em}
		\State\label{TradiionalADMM-step:4} 
		$
		X_{k+1} \gets
		\mathop{\operatorname{argmin}}\limits_{X} \|X \|_{*} +  
		\scalebox{1.3}{$
			\frac{1}{2\tau_k} 
			$}
		\|
		X + E_{k} - M - \tau_{k} \cdot Y_{k}
		\|_{F}^{2} \equiv D_{\tau_k}(M - E_k + \tau_k \cdot Y_k)
		$
		~\vspace{0.2em}
		\State\label{TradiionalADMM-step:5}   $
		E_{k+1} \gets G_{\bar{\Omega}}(M - X_{k+1} + \tau_{k} \cdot  Y_{k+1}) 
		$
		~\vspace{0.2em}
		\State $
		Y_{k+1} \gets Y_{k} +   
		\scalebox{1.3}{$\frac{1}{\tau_k}$}  \cdot (M - X_{k+1} - E_{k+1} )
		$
		~\vspace{0.2em}
		\State $\tau_{k+1} \gets   
		\max\big(\alpha \cdot
		\tau_k, \tau_\mathrm{min} 
		\big)
		$
		and $k \gets k+1$
		\EndWhile
		~\vspace{0.2em}
		\State  $X_\mathit{COMP} \gets X_{k}$
		\EndProcedure
	\end{algorithmic}
	\label{alg:ADMM}
\end{algorithm}

\subsection{Generalization of Matrix Completion over Higher-Order Generalized Scalars}
Following the application of ADMM to the optimization of Equation \ref{equation:convex}, many authors have proposed methods to extend the completion process to third-order arrays. For example, using Kilmer et al.'s ``t-product'' model, Lu et al. extended the above completion approach to third-order tensors~\cite{lu2018exact}.

Although called tensor algorithms, Lu et al.'s approach \cite{lu2018exact} and other variants \cite{Zhang2014Novel,xue2018low,lu2019tensor} are essentially matrix completion algorithms operating on generalized first-order scalars. However, as noted above, the order of generalized scalars can actually be defined as higher. 
Since higher-order arrays encapsulate more structural information than their lower-order counterparts in real-world scenarios, we exploit this aspect by increasing the order of the arrays via a pixel neighborhood strategy originally introduced in \cite{liao2020general} but largely overlooked by the research community.

\begin{figure}[!htb]
\includegraphics{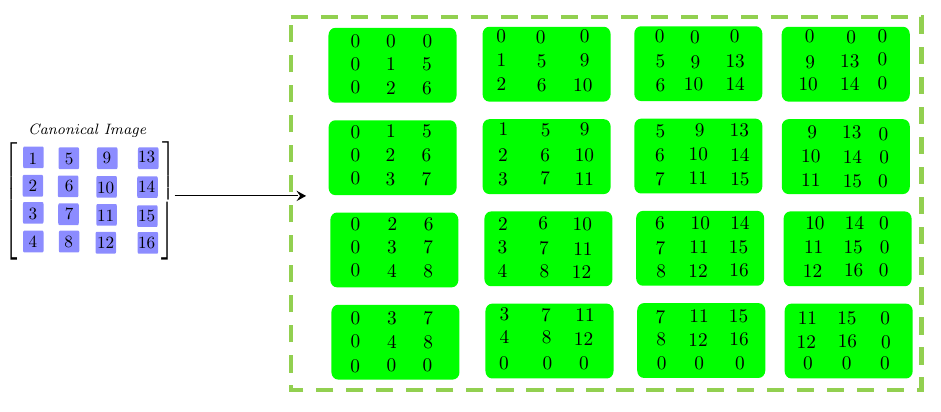}
\caption{Elevating a $4\times 4$ grayscale image to a fourth-order array using a central $3\times 3$ pixel neighborhood strategy.}
\label{fig:neighborhood-strategy}
\end{figure}

Specifically 
Figure \ref{fig:neighborhood-strategy} shows the application of a ``$3\times 3$ pixel neighborhood'' strategy to increase the order of a $4\times 4$ pixel grayscale image. Note that the figure represents the result of the order-four array as a two-dimensional array of two-dimensional blocks.

The use of this pixel neighborhood strategy results in a four-order array of size $3\times 3\times 4\times 4$, which is interpreted as a $4\times 4$ matrix with 
generalized scalars (i.e., t-scalars) of size $3\times 3$.

Note also that someone might prefer the underlying array format in the form of 
$4\times 4\times 3\times 3$ instead of $3\times 3\times 4\times 4$. We call the $3\times 3\times 4\times 4$ construct the little-endian array format of generalized matrices, and the other the big-endian format of generalized matrices.

The endian protocol is not an integral part of the algebraic definitions of t-matrices. Several array programming languages, including MATLAB and Python (NumPy), provide convenient tools to facilitate conversion between endian protocols. While Kilmer et al.'s ``t-product'' model uses big-endian, we use little-endian, which has the advantage of aligning the multimode multiplication formulation of the Fourier transform of generalized matrices (as in Equation \ref{equation:89785656}) with that of generalized scalars (as in Equation \ref{equation:89675656}).

Since each pixel value can be elevated to a generalized scalar (i.e., t-scalar), allowing the conversion of a traditional matrix into a generalized one with higher-order fixed-size arrays as entries, the extension of algorithm \ref{alg:ADMM} for generalized matrix completion is straightforward. The generalization of line \ref{TradiionalADMM-step:4} in Algorithm \ref{alg:ADMM} can be achieved using TSVD, as described in Algorithm \ref{alg:TensorSVT}. Meanwhile, line \ref{TradiionalADMM-step:5} in Algorithm \ref{alg:ADMM} is extended by the elevated linear operator $G_{\bar{\Theta}}$, which will 
which preserves entries within the enhanced set $\bar{\Theta}$ and sets those outside $\bar{\Theta}$ to the zero t-scalar $\dot{z}$.

We propose a generalized matrix (t-matrix) completion algorithm for recovering multispectral images with missing values, as described in Algorithm \ref{alg:generalized-ADMM}, where $M \in \mathbbm{R}^{D_1\times D_2\times D_3}$, $\Omega$ represents a random non-empty proper subset of the Cartesian product $[D_1]\times [D_2]\times [D_3]$, the result $\Theta$ is a proper subset of $[I_1]\times [I_2]\times [D_3]\times [D_1]\times [D_2]$, and the size of t-scalars is $I_1\times I_2\times D_3$, where $I_1,I_2$ are both odd numbers.
\begin{algorithm}[!htb]
	\caption{Higher Order TNN: ADMM for recovering an image with missing values}
	\begin{algorithmic}[1]
		\Procedure{$X_\mathit{COMP}$ = SpectralImageCompletion}{$M, \Omega$} 
		~\vspace{0.2em}
		\State Assign an improbable value, such as $-1$, to the missing entries indexed by $\Omega$.
		~\vspace{0.2em}
		\State\label{state:line002} 
		Construct $M_1, M_2,\dots, M_{D_3} \in \mathbbm{R}^{I_1\times I_2\times D_1\times D_2}$ using the $I_1\times I_2$ 
		neighborhood strat- 
		\Statex \hspace{1.7em}egy for every frontal slice of $M$. ~\vspace{0.2em}
		\State\label{state:line003}
		Construct $M_\mathit{UP} \in \mathbbm{R}^{I_1\times I_2 \times D_1\times D_2\times D_3}$ by aligning $M_1, M_2,\dots,M_{D_3}$ along the mode-5
		\State
		\label{state:line004} 
		\vspace{0.2em}
		Convert $M_\mathit{UP}$ into a generalized matrix $\dot{M} \in C^{D_1\times D_2} \equiv \mathbbm{R}^{I_1\times I_2 \times D_3 \times D_1\times D_2}$ by 
		\Statex \hspace{1.7em}permuting of the indices of array.
		~\vspace{0.2em}
		\State Store the positions of the entries ``$-1$'' of $\dot{M}$ within $\Theta$.
		~\vspace{0.2em}
		\State Initialization: $k \gets 0$,  $\dot{Y}_0 \gets ~\text{array of zeros}$,  
		$\dot{E}_0 \gets ~\text{array of zeros}$, 
		$\alpha \gets 0.9$, 
		\Statex \hspace{1.5em}$\tau_\mathrm{min} \gets 10^{-6}$  ~\vspace{0.2em}
		\While {neither convergence nor predefine maximum iterations reached}~\vspace{0.2em}
		\State\label{step:4444} 
		$
		\dot{X}_{k+1} \gets \mathrm{TSVT}(\dot{M} - \dot{E}_k + \tau_k \cdot \dot{Y}_k, \tau_k)
		$ 
		~\vspace{0.2em}
		\State\label{step:5555}   $
		\dot{E}_{k+1} \gets G_{\bar{\Theta}}(\dot{M} - \dot{X}_{k+1} + \tau_{k} \cdot  \dot{Y}_{k+1}) 
		$
		~\vspace{0.2em}
		\State $
		\dot{Y}_{k+1} \gets Y_{k} +   
		\scalebox{1.3}{$\frac{1}{\tau_k}$} (\dot{M} - \dot{X}_{k+1} - \dot{E}_{k+1} )
		$
		~\vspace{0.2em}
		\State $\tau_{k+1} \gets  
		\max\big(\alpha \cdot 
		\tau_k, \tau_\mathrm{min} 
		\big)
		$
		and $k \gets k+1$
		\EndWhile
		~\vspace{0.2em}
		\State\label{state:line014}  
		Use the row-index-first (MATLAB compliant) protocol to reshape $\dot{X}_k$ into an ~\vspace{0.2em}
		\Statex \hspace{1.7em}$I_1I_2 \times D_3D_1D_2$  
		matrix, extract the central row, and subsequently reshape it with ~\vspace{0.2em}
		\Statex \hspace{1.7em}the row-index-first protocal  
		into an array
		$X_\mathit{DOWN} \in \mathbbm{R}^{D_3\times D_1\times D_3}$.
		\State\label{state:line015}   
		Permute the indices of $X_\mathit{DOWN}$ to convert it into an array in $\mathbbm{R}^{D_1\times D_3\times D_3}$. ~\vspace{0.2em}
		\State Adjust the entries of $X_\mathit{DOWN}$ to nonnegative integers and store the adjusted ~\vspace{0.2em}
		\Statex \hspace{1.7em}array $X_\mathit{DOWN}$ as $X_\mathit{COMP}$, as the recovered multispectral image.
		\EndProcedure
	\end{algorithmic}
	\label{alg:generalized-ADMM}
\end{algorithm}

The lines \ref{state:line002}, \ref{state:line003} and \ref{state:line004} of the 
of the proposed algorithm
up-convert the input multispectral image $M$, an initial array $D_1\times D_2\times D_3$, into a $D_1\times D_2$ t-matrix $M_\mathit{UP}$, with t-scalars of size $I_1\times I_2\times D_3$. Conversely, line \ref{state:line014} down-converts the optimal generalized matrix $\dot{X}_{k}$ into a third-order array $X_\mathit{DOWN}$.

Algorithm \ref{alg:generalized-ADMM} is based on the tensor completion algorithm of Lu et al~\cite{lu2018exact}. Section \ref{section:experiments} of this paper focuses primarily on its empirical validation.

\section{Rank Considerations}
\label{section:RankConsiderations}
Matrix completion aims to recover a complete low-rank matrix. Its generalization aims to recover an analogous higher-order t-matrix. However, the rank of a t-matrix is not yet specified, so let's look at some novel rank notions defined for higher order arrays.

\subsection{Tubal Rank and Average Rank}

An RGB image, a special case of multispectral images, consists of three monochromatic channels. Each channel in real RGB images can be adequately approximated by lower-rank matrices. However, when viewed as a third-order tensor, the canonical rank of an RGB image, which is defined by the minimal set of rank-one tensor addends, becomes computationally intractable. Consequently, the canonical tensor rank is unsuitable for modeling the optimal recovery of an RGB image.

Kilmer et al.'s approach is to introduce a novel rank concept, called tubal rank, for a third-order array. Specifically, for a given third-order array $\dot{X}$, with its TSVD defined as $\dot{X} = \dot{U} \circ \dot{S} \circ \dot{V}^{*}$, the tubal rank of $\dot{X}$ corresponds to the number of non-zero (i.e., not equal to 
$\dot{z}$) diagonal t-scalars in $\dot{S}$.
Nevertheless, by this definition, a t-matrix of full tubal rank can consist of a full-rank matrix as one of its spectral slices, with all other spectral slices being zero matrices.

To address this problem, Lu et al. proposed to define the average of all spectral slice ranks as the ``average rank'' of a generalized matrix~\cite{lu2019tensor}. This ``average rank'' definition is more appropriate than the tubal rank.  However, this term is only used for Lu et al.'s 
generalization of robust component analysis~\cite{lu2019tensor}, not for the higher-order array recovery problem presented in~\cite{lu2018exact}.
Moreover, despite the potential for ``average rank'' to be fractional, the mathematical justification for ``average rank'' has not been adequately addressed.

\subsection{Higher-Order Rank and Its Trace Variant}
\label{section:Higher-OrderRankandItsTraceVariant}
In addition to the tubal rank of Kilmer et al. and the average rank of Lu et al., another relevant concept is the higher-order rank introduced by Liao and Maybank in their paper \cite{liao2020generalized}. Specifically, given a t-matrix $\dot{X}$ with its tensor singular value decomposition (TSVD) $\dot{X} = \dot{U} \circ \dot{S} \circ \dot{V}^{*}$, the higher-order rank of $\dot{X}$ is a non-negative t-scalar. It is computed as the sum of the diagonal entries of the product $\dot{S}^{\dagger} \circ \dot{S}$, where $\dot{S}^{\dagger}$ denotes the pseudoinverse of $\dot{S}$.

The above definition corresponds to the analogous concept for traditional matrices. The pseudo-inverse of a t-matrix can be computed using spectral slices, similar to Algorithms \ref{alg:TensorSVD} and \ref{alg:TensorSVT}.
Specifically, the pseudo-inverse $\dot{X}^{\dagger}$ of a t-matrix 
is defined by assigning the pseudo-inverse of each spectral slice to its corresponding slice in the result.
It is not difficult to verify that the pseudo-inverse $\dot{X}^{\dagger}$ defined above is equal to the product $\dot{V} \circ \dot{S}^{\dagger} \circ \dot{U}^{*}$. That is, the equation $\dot{X}^{\dagger} = \dot{V} \circ \dot{S}^{\dagger} \circ \dot{U}^{*}$ holds for any t-matrix $\dot{X}$.

From the previous definition, it is easy to see that the higher-order rank of any t-matrix is a nonnegative t-scalar. It can be sorted alongside comparable nonnegative counterparts using the partial order introduced in Section \ref{section:GeneralizedScalars}.
However, sometimes we prefer a more efficient rank notion similar to the fully ordered ones proposed by Kilmer et al. and Lu et al. as opposed to the partially ordered higher order rank.
Reassuringly, the Szpilrajn extension theorem asserts that the partially ordered rank system proposed in \cite{liao2020generalized} can always be extended to a fully ordered construct.

There are several strategies for transforming the higher order rank system into its fully ordered equivalents. Considering any higher-order rank of a t-matrix, all spectral points (i.e., Fourier entries) are nonnegative integers. Consequently, Kilmer et al.'s tubal rank denotes the maximum value among these spectral points, while Lu et al.'s average rank is equal to their arithmetic mean.

Typically, in most scenarios, the average rank of Lu et al. is considered a superior statistic for a higher order rank.
However, to avoid fractional rank values, we propose to use the sum, rather than the arithmetic mean, of the spectral points of a higher-order rank to define its corresponding fully ordered rank.
Furthermore, since every t-scalar also functions as a finite-dimensional linear endomorphic operator, the previously defined ``sum rank'' of a t-matrix is equivalent to the trace of the higher-order rank, so it would be appropriate to formally label it as the ``trace rank''.

The fairness of the above definitions can be given by using the representation theory known in the mathematical community. We give a brief discussion of representation theory with its application to justify the above definition in  Appendix.

The fairness of the above definitions can be given by using the representation theory known in the mathematical community. We give a brief discussion of representation theory with its application to justify the above definition in  Appendix \ref{appendix:appendix}.

\section{Experiments}
\label{section:experiments}

This part presents experimental validation and performance analysis of the related algorithms.

\subsection{Experiments on Simulated Random Data}
To evaluate the completion capability of the proposed higher-order TNN algorithm, we use simulated random t-matrices for verification. Specifically, we generate two random t-matrices $\dot{P} \in C^{D\times r} $ and $\dot{Q} \in C^{r\times D} $, which are represented as arrays in $\mathbbm{R}^{I_1\times I_2 \times I_3\times D\times r}$ and $\mathbbm{R}^{I_1\times I_2
	\times I_3 \times r\times D}$, where $I_1\times I_2 \times I_3 = 3\times 3\times 3$ and $r < D$.
The parameter $r$ is (with high probability) the tubal rank as defined by Kilmer et al. at~\cite{kilmer2011factorization,kilmer2013third}. The real numbers in the underlying arrays of $\dot{P}$ and $\dot{Q}$ are independently sampled from the distribution $\mathcal{N}(0, 1)$.

The product $\dot{Y} = \dot{P} \cdot \dot{Q}$ gives a random t-matrix in $C^{D\times D} \equiv \mathbbm{R}^{I_1\times I_2 \times I_3 \times D \times D}$. The trace rank of $\dot{Y}$ is, with high probability, $\operatorname{rank}_\mathrm{trace} \dot{Y} = I_1I_2I_3 \cdot r$. From the underlying array of $\dot{Y}$, we uniformly select entries to simulate missing data. The resulting incomplete $\dot{Y}$, with a varying percentage of missing entries and a rank parameter $r$ for $\dot{Y}$, serves as input to the Higher-order TNN algorithm.

The Higher-order TNN algorithm produces a t-matrix $\dot{X} \in C^{D\times D} \equiv \mathbbm{R}^{I_1\times I_2\times I_3 \times D\times D}$, which serves as an estimated version of $\dot{Y}$. If $\
\mathit{RSE} \doteq {\|\operatorname{tensor}(\dot{Y}) - \operatorname{tensor}(\dot{X})\|_F}/{\| \operatorname{tensor}(\dot{Y})\|_F}$ is less than a threshold, the completion by the Higher-order TNN algorithm is considered successful.

\begin{figure}[!htb]
	\centering
	\small
	\begin{minipage}{0.99\textwidth}
		\centering
		\begin{tabular}{rl}
			\includegraphics[height=16.2em]{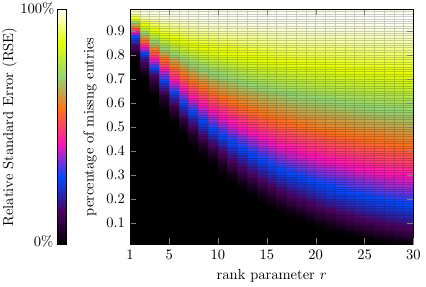}	&
			\raisebox{0.4em}{\includegraphics[height=15.5em]{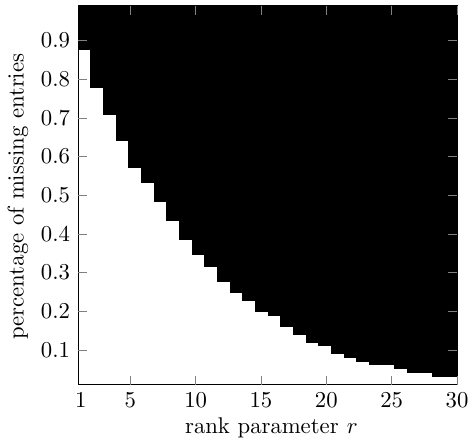}} \\
			\multicolumn{1}{c}{\hspace{6em}\makecell{(a) RES distribution \\ $D = 40$}} &  \multicolumn{1}{c}{\hspace{1.5em} 
				\makecell{(b) phase transition \\ $D = 40$}}
		\end{tabular}
	\end{minipage}
	
	\begin{minipage}{0.99\textwidth}
		\centering	
		\begin{tabular}{rl}
			\includegraphics[height=16.2em]{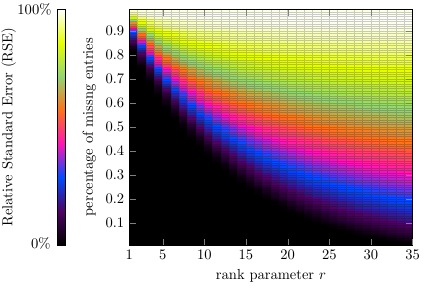}	&
			\raisebox{0.4em}{\includegraphics[height=15.5em]{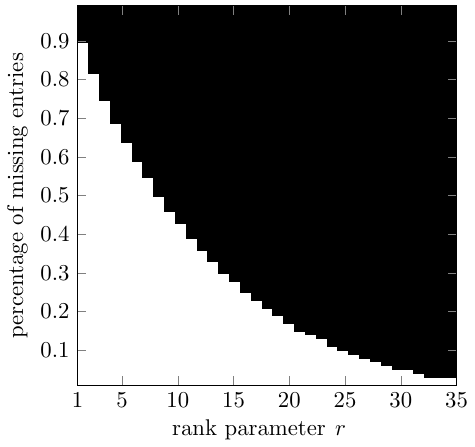}} \\
			\multicolumn{1}{c}{\hspace{6em}
				\makecell{(c) RSE distribution \\$D = 50$}} &  \multicolumn{1}{c}{\hspace{1.5em}
				\makecell{(d) phase transition \\$D = 50$}}
		\end{tabular}
	\end{minipage}
	\caption{RSE Distributions and Phase Transitions (Threshold $\mathit{RSE} = 1\times 10^{-2}$  ) of Higher Order TNN, with $D = 40$ and $50$}
	\label{fig:RSE-distribution-phase-transition}
\end{figure}

Figure \ref{fig:RSE-distribution-phase-transition} illustrates the RSE distributions and phase transitions for fifth-order array completions using the proposed higher-order TNN. Figure \ref{fig:RSE-distribution-phase-transition}(a) shows the RSE distribution with the parameter $D = 40$. Figure \ref{fig:RSE-distribution-phase-transition}(b) shows a phase transition corresponding to $D = 50$, where the white and black dots represent successful ($\mathit{RSE} <1\times 10^{-2}$) and failed completions, respectively.
Figures \ref{fig:RSE-distribution-phase-transition}(a) and \ref{fig:RSE-distribution-phase-transition}(c) show the RSE distributions with $D = 40$ and $D = 50$ 
Figures \ref{fig:RSE-distribution-phase-transition}(b) and \ref{fig:RSE-distribution-phase-transition}(d) show the phase transitions associated with $D = 40$ and $D = 50$, respectively.

\subsection{Experiments on BSD Color Images}
In the following experiments, we use the Berkeley segmentation dataset as a benchmark to compare the performance of four related algorithms: Tubal-TNN~\cite{Zhang2014Novel}, T-TNN~\cite{xue2018low}, TNN~\cite{lu2018exact}, and our Higher-order TNN.
Three RGB images, namely ``Resort'', ``Insect'', and ``Seagulls'' are selected for the first experiment. These images are represented as $321\times 481\times 3$ unsigned integer arrays.

To compare the completion performance, we randomly select $70\%$ of the pixel values of each image as ``missing'' entries. The uncompleted observed images, with missing values set to zero, provide a visual representation in Figure \ref{figure:resort-insect-gulls}, which shows the original complete images alongside their incomplete versions.

\begin{figure}[!htb]
\vspace{1em}
\centering	
\begin{tabular}{c}
\includegraphics[width=0.3\textwidth]{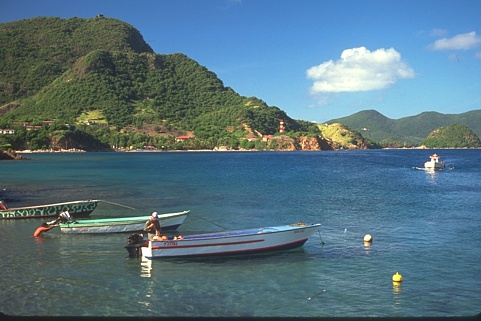} 
\includegraphics[width=0.3\textwidth]{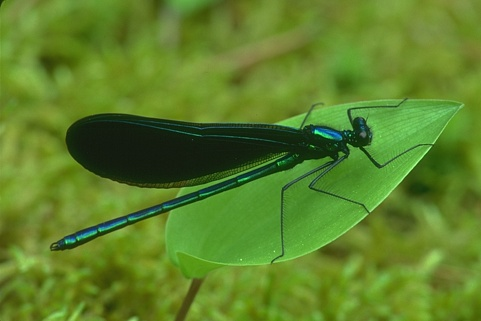} 
\includegraphics[width=0.3\textwidth]{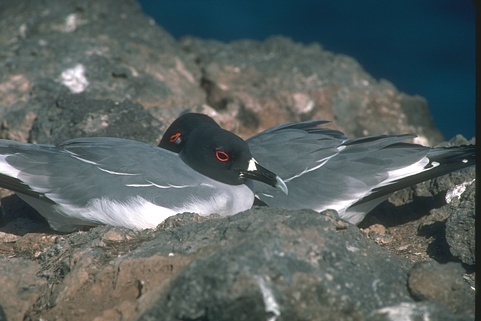} \\
(a) Original images \vspace{0.3em}\\
\includegraphics[width=0.3\textwidth]{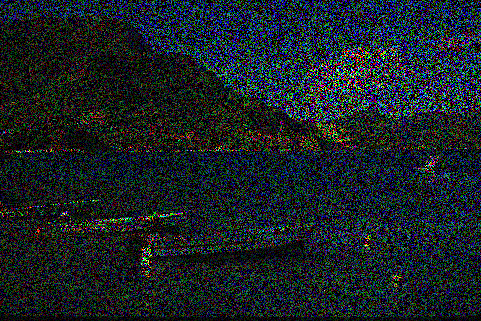} 
\includegraphics[width=0.3\textwidth]{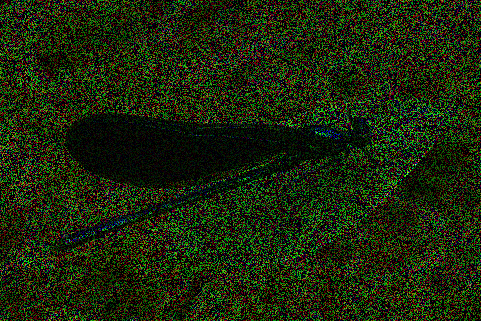} 
\includegraphics[width=0.3\textwidth]{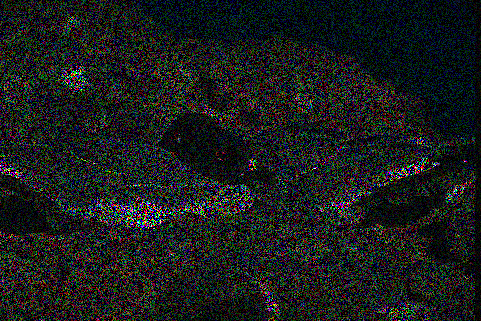} \\
(b) Observed images with missing $70\%$ entries 
\vspace{0.3em}\\
\end{tabular}
\caption{
Three original and observed images: ``Resort'', ``Insect'', and ``Seagulls'' from the Berkeley Segmentation Dataset.
}
\label{figure:resort-insect-gulls}
\end{figure}

We use the three competing algorithms and our Higher-Order TNN, described in Algorithm \ref{alg:generalized-ADMM}, to obtain an optimal, complete RGB image of equal size. The quality of the image completion is quantified by the Peak Signal to Noise Ratio (PSNR), defined as
\begin{equation}
	\mathit{PSNR} = 10 \cdot \log_{10} \frac{D_1 \cdot D_2 \cdot D_3}{ \| X_\mathit{COMP} - M \|_F^{2}}
\end{equation}
where $D_1 \times D_2 \times D_3$ is the size of the RGB image. Note that before computing the PSNR, the values of both $X_\mathit{COMP}$ and $M$ are normalized to $1$, thus adjusting the ``peak'' value involved in the PSNR computation from $255$ to $1$.

Figure \ref{figure:resort-insect-gulls} presents visual and quantitative comparisons of the performance of four competing algorithms in completing the observed images: ``Resort'', ``Insect'', and ``Seagulls'' from the Berkeley Segmentation Dataset.
The proposed Higher-Order TNN outperforms its competitors in terms of PSNRs by at least $1$ dB, $1.4$ dB, and $1.6$ dB, respectively.

\begin{figure}[!htb]
	\centering	
	\begin{tabular}{c}
		\includegraphics[width=0.3\textwidth]{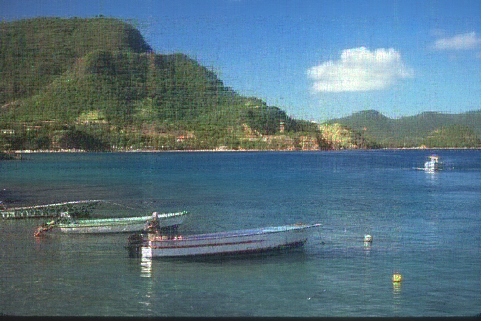} 
		\includegraphics[width=0.3\textwidth]{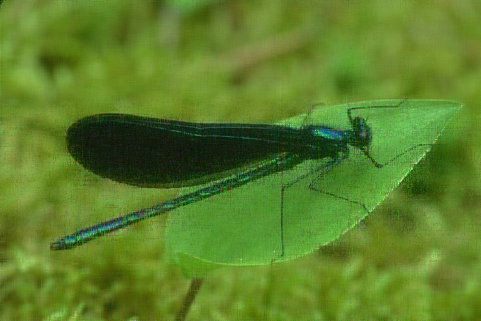} 
		\includegraphics[width=0.3\textwidth]{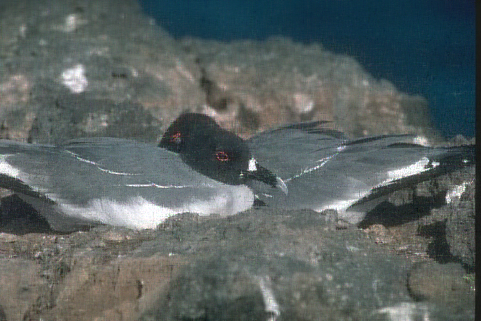} \\
		(a) Tubal-NN~\cite{Zhang2014Novel}\vspace{0.3em}\\
		
		\includegraphics[width=0.3\textwidth]{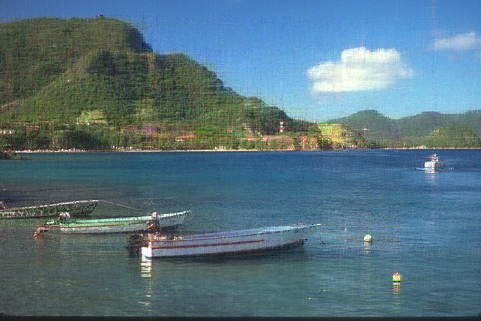} 
		\includegraphics[width=0.3\textwidth]{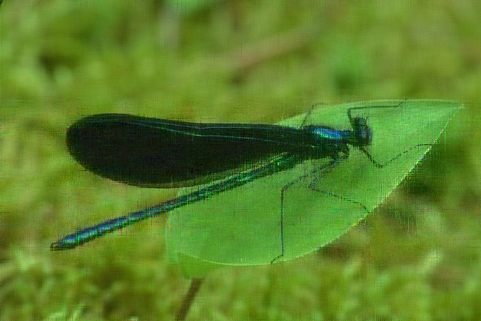} 
		\includegraphics[width=0.3\textwidth]{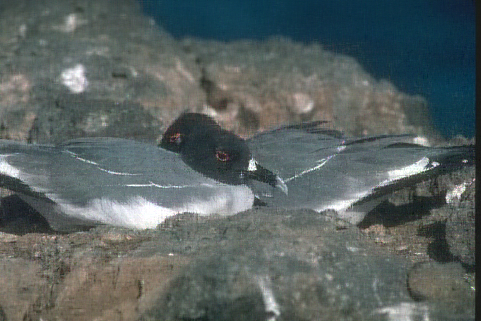} \\
		(b) T-TNN~\cite{xue2018low}\vspace{0.3em}\\
		
		\includegraphics[width=0.3\textwidth]{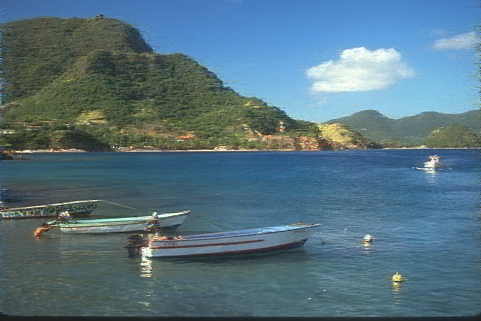} 
		\includegraphics[width=0.3\textwidth]{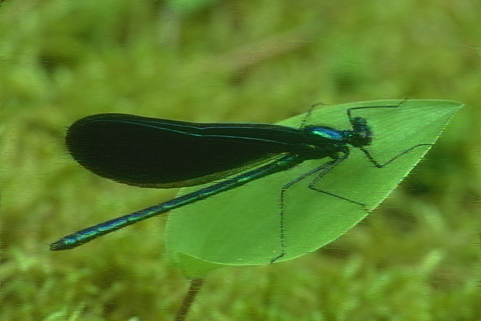} 
		\includegraphics[width=0.3\textwidth]{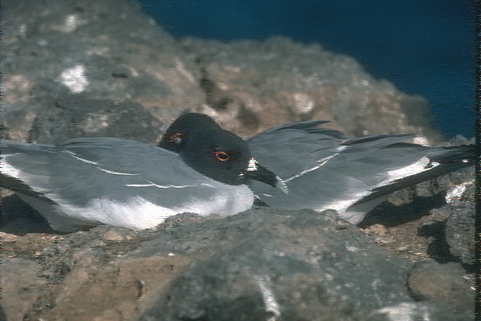} \\
		(c) TNN~\cite{lu2018exact}\vspace{0.3em}\\
		
		\includegraphics[width=0.3\textwidth]{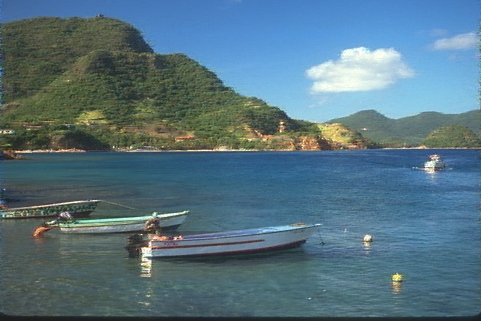} 
		\includegraphics[width=0.3\textwidth]{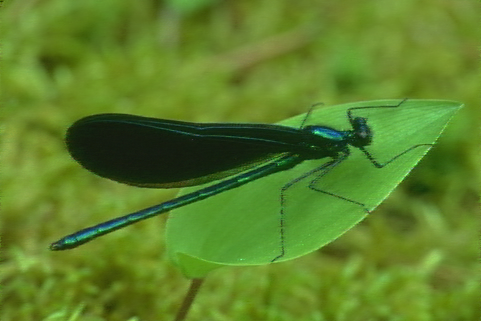} 
		\includegraphics[width=0.3\textwidth]{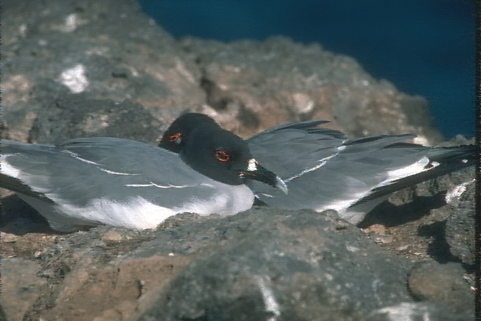} \\
		(d) Higher-Order TNN (ours) \vspace{0.3em}\\		
		
	\end{tabular}
	
\vspace{0.5em}
{\centering PSNR comparison of four competing algorithms}
\hspace{-2em}\includegraphics[width=0.9\textwidth]{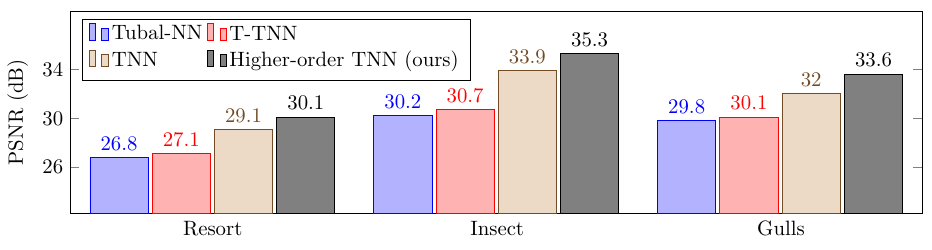} \\
\caption{
Visual and quantitative comparisons of the performance of four related algorithms in completing the images ``Resort'', ``Insect'', and ``Seagulls''
}
\label{figure:resort-insect-gulls}
\end{figure}

In our second experiment, we use three different images --- ``Temple'', ``Chapel'', and ``Grass-flower'', each with $50\%$ entries randomly missing, to perform completions analogous to the first experiment on the Birkeley images, each with $70\%$ entries randomly missing.
Figure \ref{fig:temple-chapel-grass-flower} shows the original images along with their incomplete versions.

Figure \ref{fig:temple-chapel-grass-flower} shows the visual and quantitative comparisons of the four related image completion algorithms. Consistent with the results of the first experiment, the higher-order TNN significantly outperforms the other algorithms, achieving gains of at least $1.5$ dB, $1.2$ dB, and $2.2$ dB, respectively.

\begin{figure}[!htb]
	\centering	
	\begin{tabular}{c}
		\includegraphics[width=0.3\textwidth]{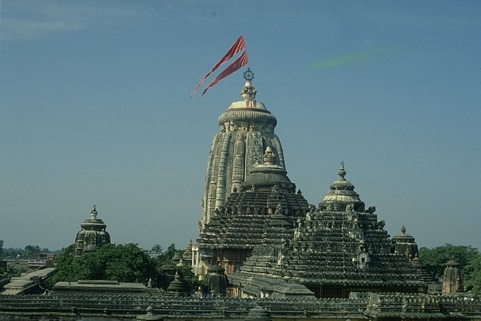} 
		\includegraphics[width=0.3\textwidth]{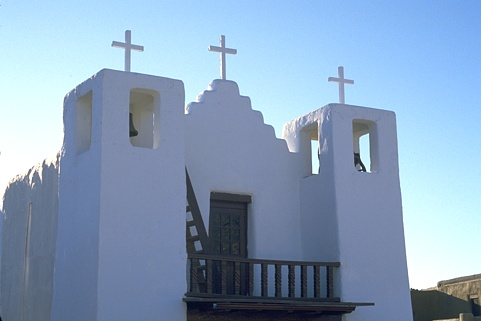} 
		\includegraphics[width=0.3\textwidth]{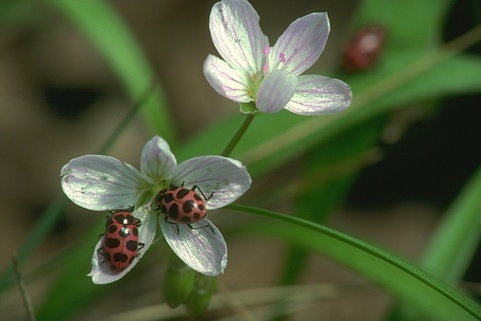} \\
		(a) Orignal images \vspace{0.3em}\\
		\includegraphics[width=0.3\textwidth]{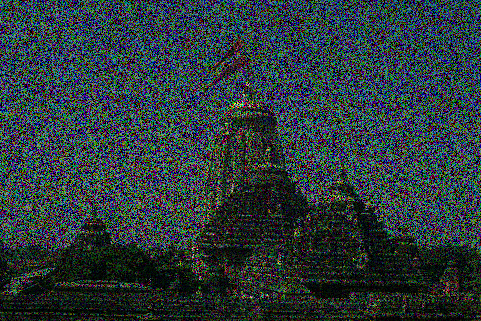} 
		\includegraphics[width=0.3\textwidth]{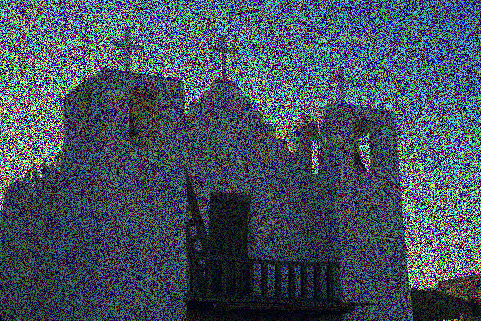} 
		\includegraphics[width=0.3\textwidth]{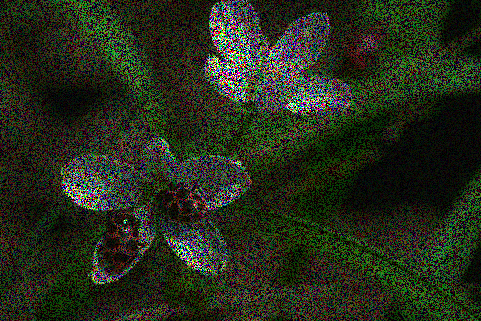} \\
		(b) Observed images with $50\%$ entries missing 
		\vspace{0.3em}\\
	\end{tabular}
\caption{
	Three original and observed images: ``Temple'', ``Chapel'', and ``Grass-flower'' from the Berkeley Segmentation Dataset.
}
\label{fig:temple-chapel-grass-flower}
\end{figure}

\begin{figure}[!htb]
	\centering	
	\begin{tabular}{c}
		\includegraphics[width=0.3\textwidth]{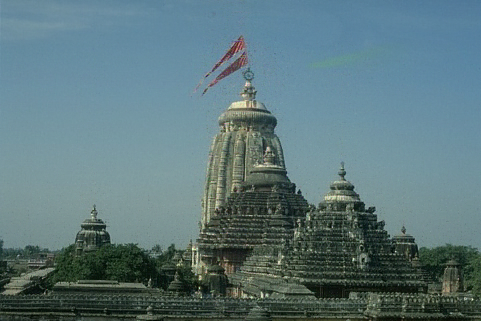} 
		\includegraphics[width=0.3\textwidth]{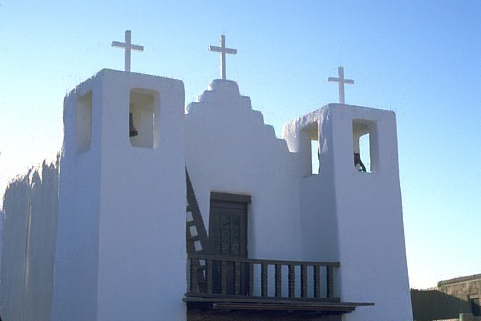} 
		\includegraphics[width=0.3\textwidth]{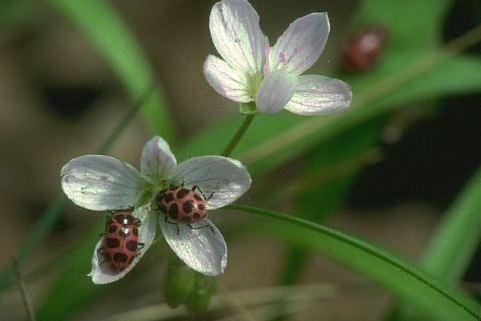} \\
		(a) Tubal-NN~\cite{Zhang2014Novel}\vspace{0.3em}\\
		
		\includegraphics[width=0.3\textwidth]{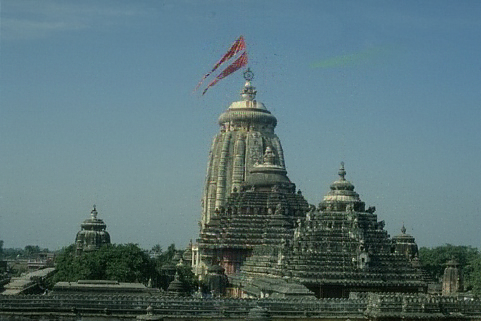} 
		\includegraphics[width=0.3\textwidth]{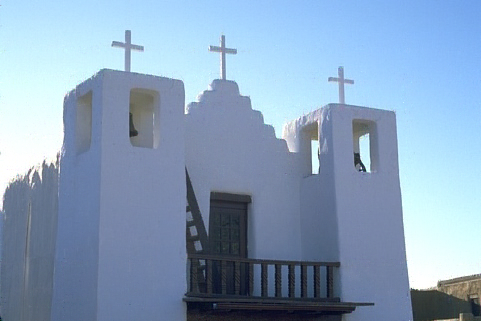} 
		\includegraphics[width=0.3\textwidth]{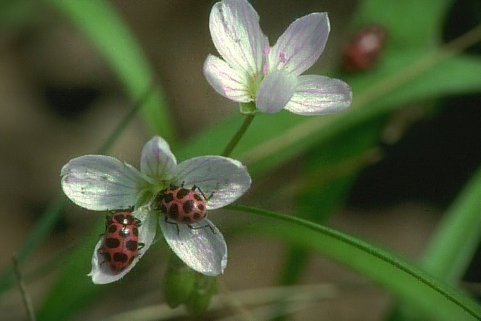} \\
		(b) T-TNN~\cite{xue2018low}\vspace{0.3em}\\
		
		\includegraphics[width=0.3\textwidth]{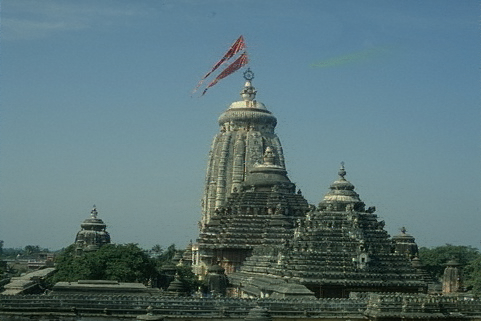} 
		\includegraphics[width=0.3\textwidth]{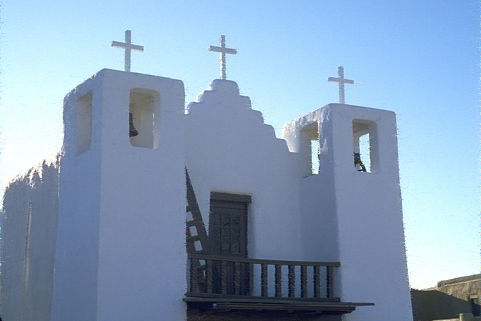} 
		\includegraphics[width=0.3\textwidth]{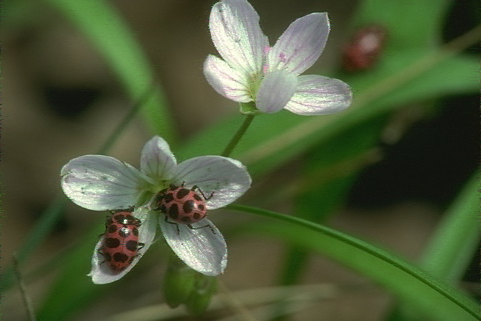} \\
		(c) TNN~\cite{lu2018exact}\vspace{0.3em}\\
		
		\includegraphics[width=0.3\textwidth]{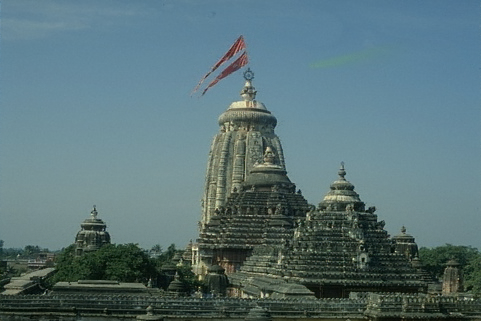} 
		\includegraphics[width=0.3\textwidth]{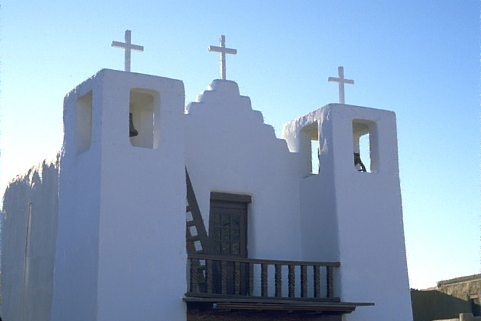} 
		\includegraphics[width=0.3\textwidth]{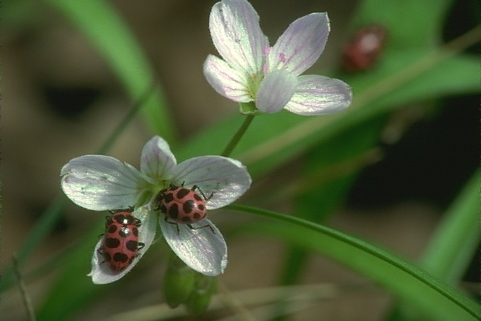} \\
		(d) Higher-Order TNN (ours) \vspace{0.3em}\\		
		
	\end{tabular}
	
	\hspace{-2em}\includegraphics[width=0.9\textwidth]{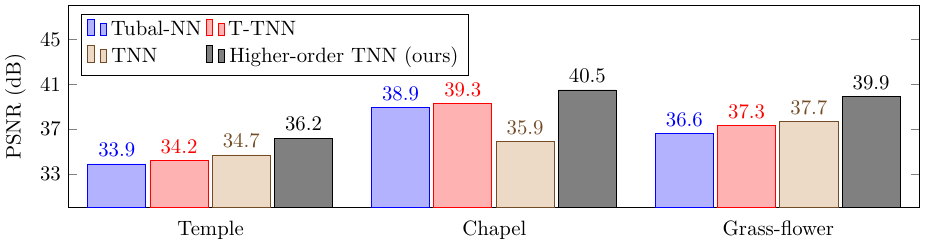} \\
	\caption{Visual and quantitative comparisons of the performance of four related algorithms in completing the images ``Temple'', ``Chapel'', and ``Grass-flower''}
\label{fig:temple-chapel-grass-flower}
\end{figure}

After experiments on $6$ RGB images, our research now includes $10$ randomly selected RGB images from the Berkeley Segmentation Dataset, with the percentage of missing entries set to increments within $\{0.1, 0.2,\dots,0.9\}$.

\begin{figure}[!htb]
	\begin{minipage}{\textwidth}
		\begin{tabular}{c}
			\includegraphics[width=0.18\textwidth]{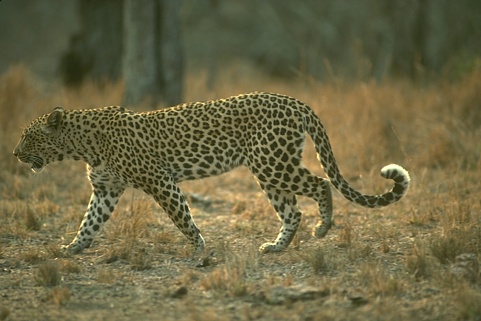}
			\includegraphics[width=0.18\textwidth]{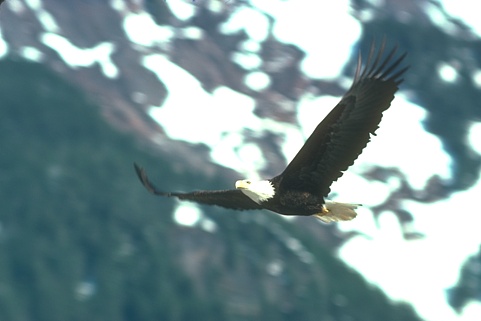}
			\includegraphics[width=0.18\textwidth]{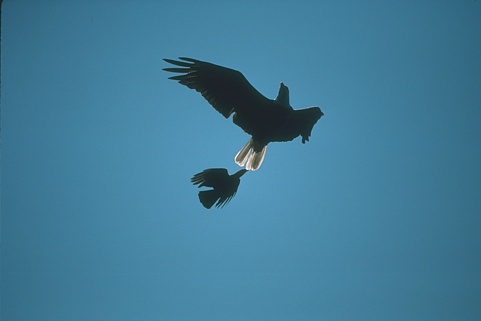}
			\includegraphics[width=0.18\textwidth]{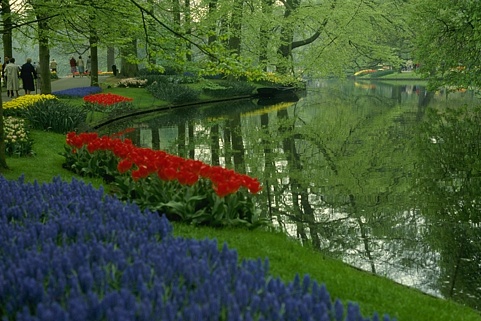}
			\includegraphics[width=0.18\textwidth]{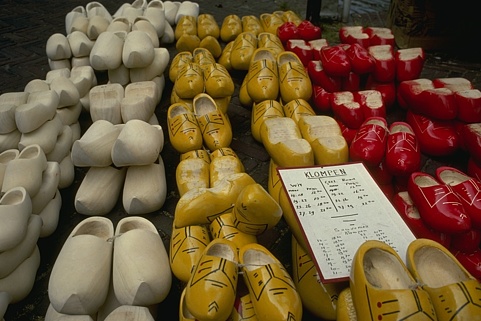} \\
			\includegraphics[width=0.18\textwidth]{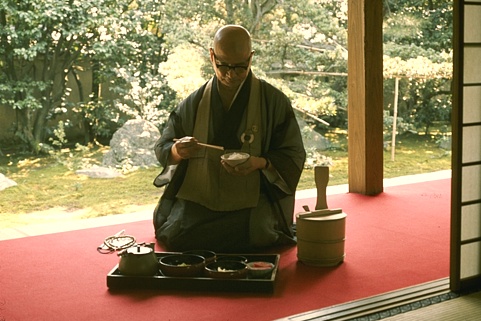}
			\includegraphics[width=0.18\textwidth]{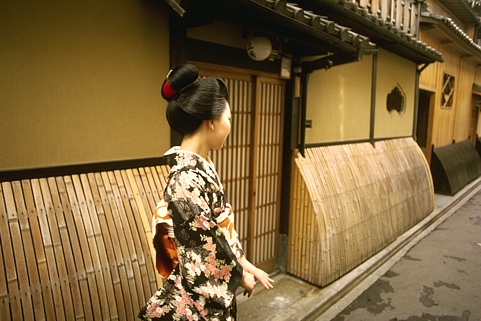}
			\includegraphics[width=0.18\textwidth]{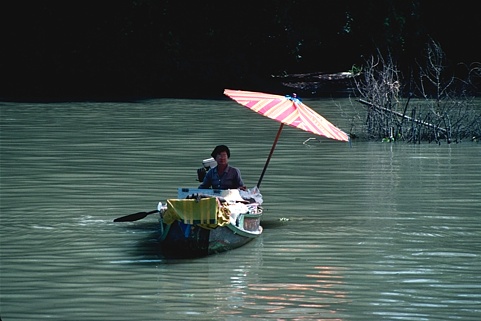}
			\includegraphics[width=0.18\textwidth]{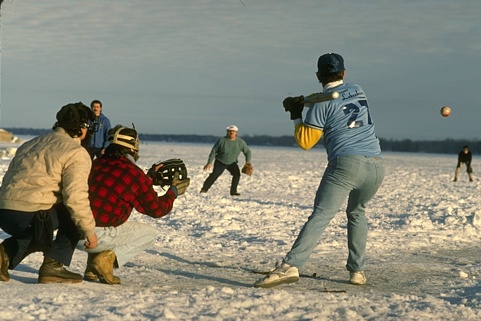}
			\includegraphics[width=0.18\textwidth]{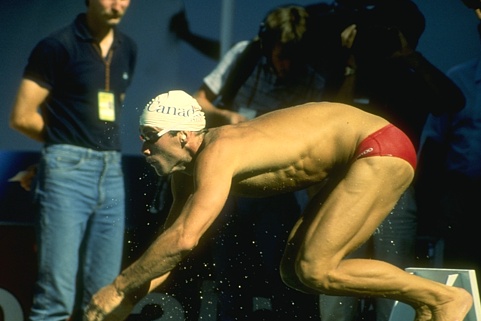} 
		\end{tabular}
		\caption{$10$ random RGB images selected from the Berkeley segmentation dataset}
		\label{fig:random-10-images}	
	\end{minipage}
	
	\vspace{1em}
	
	\begin{minipage}{\textwidth}
		\begin{tabular}{rl}	
			\includegraphics[width=0.45\textwidth]{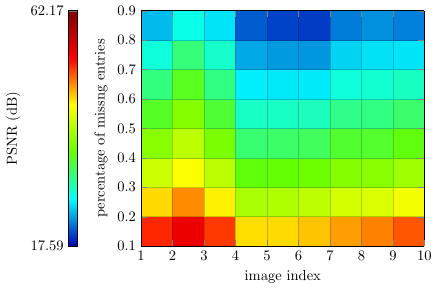} &
			\includegraphics[width=0.45\textwidth]{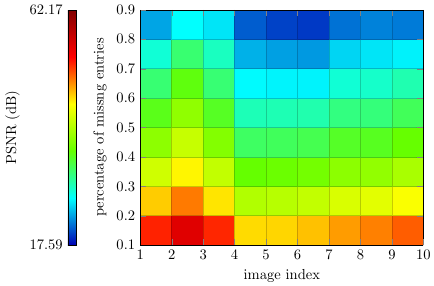} \\
			\multicolumn{1}{c}{\hspace{6.5em}(a) Tubal NN~\cite{Zhang2014Novel}} & \multicolumn{1}{c}{\hspace{6.5em}(b) T-TNN~\cite{xue2018low}} \\
			\includegraphics[width=0.45\textwidth]{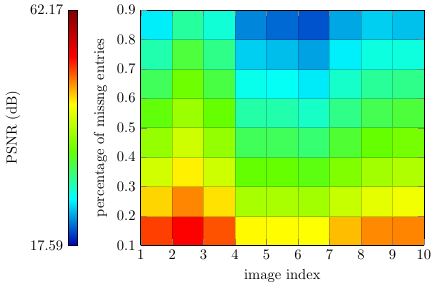} &
			\includegraphics[width=0.45\textwidth]{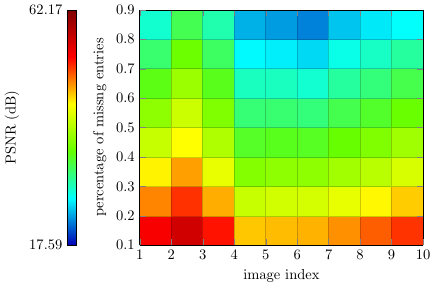} \\
			\multicolumn{1}{c}{\hspace{6.5em}(c) TNN~\cite{lu2018exact}} & \multicolumn{1}{c}{\hspace{6.5em}(d) Higher-Order TNN (ours)} \\
		\end{tabular}
		\caption{PSNR heatmaps of four completion algorithms with different percentages of missing entries on $10$ RGB images.}
		\label{fig:heatmaps}
	\end{minipage}
\end{figure}

Figure \ref{fig:random-10-images} shows the $10$ randomly selected RGB images used for the experiments. 
Figure \ref{fig:heatmaps} shows the PSNR heatmaps for four relevant algorithms: Tubal NN~\cite{Zhang2014Novel}, T-TNN~\cite{xue2018low}, TNN~\cite{lu2018exact}, and our newly developed Higher-Order TNN on the images shown in figure \ref{fig:random-10-images}. 
It's important to note that similar results were obtained in experiments on other randomly selected images not included in this paper.

These results demonstrate the superior performance of the higher-order TNN, as it outperforms its counterparts in terms of PSNR.

\begin{figure}[!htb]
	\begin{tabular}{cc}
		\includegraphics[width=0.45\textwidth]{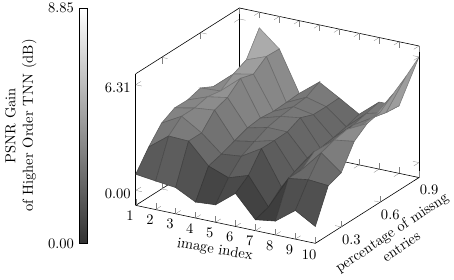} & 
		\includegraphics[width=0.45\textwidth]{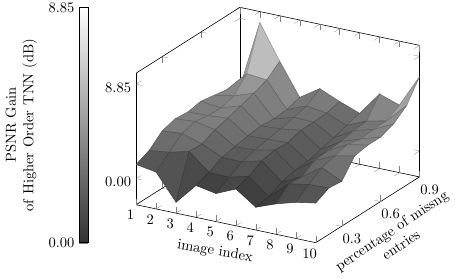} \\
		\hspace{4em}(a) PSNR gain over Tubal-NN& 
		\hspace{4em}(b) PSNR gain over T-TNN\\
		~ & 
		~ \\
		\multicolumn{2}{c}{
			\includegraphics[width=0.45\textwidth]{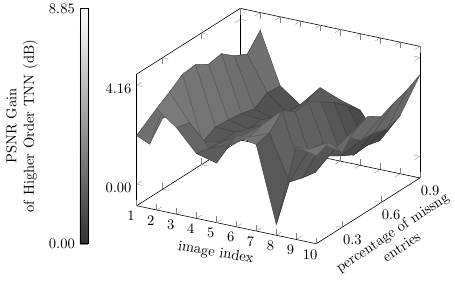}
		} \\
		\multicolumn{2}{c}{
			\hspace{4em}(c) PSNR gain over TNN
		} \\	
	\end{tabular}
	\caption{Higher order TNN's PSNR gains over its competitors.}
	\label{fig:PSNR-gain}
\end{figure}

For a clearer demonstration, Figure \ref{fig:PSNR-gain} shows the PSNR gains of our Higher-Order TNN over Tubal NN, T-TNN, and TNN. It shows that the PSNR gains of Higher-Order TNN over its counterparts are predominantly positive.

\section{Conclusions}
\label{section:Conclusions}

In this paper, we consider the problem of higher-order array completion using the higher-order t-matrix model. By adopting a consistent solution, we generalize low-rank matrix completion for the recovery of RGB images with missing entries. Our proposed ``higher-order TNN'' method outperforms competitors such as Tubal TNN, T-TNN, and TNN in terms of recovery performance, demonstrating the ability to exploit higher-order relationships within high-dimensional data and showing distinct advantages.

Our solution not only improves visual data completion, but also paves the way for broader applications. Integrating higher-order generalization into existing systems lays the foundation for more robust and efficient handling of higher-order, high-dimensional data.

We demonstrate this with a generalization of Lu et al.'s tensor completion algorithm to its higher-order version by formulating its matrix model over finite-dimensional algebra. This is achieved through a novel pixel neighborhood strategy.

The study also provides a consistent methodology for exploring various properties of the t-matrix model, including the notions of rank, norm, and inner product. Compared to the existing ``t-product'' model, our approach offers new insights into generalized scalars and matrices from a fresh perspective of representation and operator theory. Moreover, the higher-order Lagrange multipliers with generalized matrix variables add to our contributions.

The inclusion of the novel ``trace rank'', nuclear norm, Schatten $p$-norm, and the adaptation of the recent tensor completion algorithm by Lu et al. for higher-order scenarios highlight our results. The public image experiments emphasize the competitive advantage of our higher-order matrix completion algorithm in RGB image recovery.

\appendix
\section{
	Appendix: 
	A Methematical Justification}
\label{appendix:appendix}
The ADMM optimization for low-rank matrix completion described in Algorithm \ref{alg:ADMM} relies on the Lagrange multiplier given in Equation \ref{equation:lagrange}.
However, the generalized ADMM optimization presented in Algorithm \ref{alg:generalized-ADMM} currently lacks a corresponding Lagrange multiplier. To construct a valid Lagrange multiplier, we need to define a new nuclear norm, a real-valued inner product, and a Frobenius-like norm. 
While some researchers have proposed certain definitions, these are typically problem-specific and may lead to confusion or contradiction.

\subsection[\appendixname~\thesubsection]{Matrix representation for t-scalars and higher-order measures}

In the appendix, we offer an organized exposition of t-matrices via representation theory, an area that has yet to gain widespread recognition in computer science. A representation of an algebra $C$ requires a vector space $V$ and a homomorphism from $C$ into $\operatorname{End}(V)$, the endomorphism algebra of $V$.

The representation of the algebra $C$ allows to represent each t-scalar in $C\equiv \mathbbm{C}^{I_1\times \cdots \times I_N}$ as a diagonal complex matrix. The diagonal entries of the matrix correspond to the Fourier components of the t-scalar, resulting in the following mapping for all $\dot{x} \in C$:
\begin{equation}
	\dot{x} \mapsto M(\dot{x}) \doteq \operatorname{diag} \big\{
	F_1(\dot{x}),\dots,F_K(\dot{x})
	\big\} \;.
	\label{equationn:matrix-representation-tscallars}
\end{equation}
Here, $F_1(\dot{x}),\dots,F_K(\dot{x})$ represent the Fourier entries of $F(\dot{x})$.

It is obvious that $F_1(\dot{x}),\dots,F_K(\dot{x})$ are eigenvalues of both the t-scalar $\dot{x}$ and the matrix $M(\dot{x})$. The conjugate $\dot{x}^{}$ maps to the conjugate transpose of $M(\dot{x})$, giving the following one-to-one mapping for all $\dot{x} \in C$, 
\begin{equation}
	\dot{x}^{*} \mapsto  M(\dot{x})^{H} = \operatorname{diag} \left\{
	\overline{F_1(\dot{x})},\dots, \overline{F_K(\dot{x})}
	\right\}. 
\end{equation}

Since any t-scalar $\dot{x}$ is a normal operator, i.e., $\dot{x}^{*} \circ \dot{x} = \dot{x} \circ \dot{x}^{*}$, there exists a unique non-negative square root $|\dot{x} | \doteq \sqrt{\dot{x}^{*} \circ \dot{x}}$, which leads to the following one-to-one mapping:
\begin{equation}
	|\dot{x}|  
	\mapsto
	\operatorname{diag} \left\{
	|F_1(\dot{x})|,\dots, |F_K(\dot{x})|
	\right\}.
\end{equation}

In operator theory, such a non-negative square root is called a positive operator, alternatively a ``non-negative operator'', by Definition \ref{definition:89785656}, despite its less common usage. This non-negative operator (t-scalar) can appropriately be called the higher-order absolute value of $\dot{x}$.

In addition, a t-scalar behaves as an endomorphism within the finite-dimensional algebra $C$ and as such has a trace. The trace of any given t-scalar $\dot{x}$ can be computed via 
\begin{equation}  
	\operatorname{trace} (\dot{x}) = F_1(\dot{x}) +\dots + F_K(\dot{x})\;.
\end{equation}

This trace, when pertaining to a nonnegative t-scalar (a nonnegative operator), is a nonnegative real number. 
This property elevates the partial order of nonnegative t-scalars to a total order. Therefore, the trace of the higher-order absolute value $\sqrt{\dot{x}^{*} \circ \dot{x}}$, or the trace norm of $\dot{x} \in C$, can be determined.

The trace norm of a t-scalar is the nuclear norm of its matrix representation. For any t-scalar $\dot{x}$, this relation can be expressed as
\begin{equation}
	\operatorname{trace} 
	|\dot{x}|
	\doteq \operatorname{trace} M\left(
	|\dot{x}|
	\right) 
	=
	\| M(\dot{x}) \|_* ,\;\forall \dot{x} \in C. 
	\label{equation:trace-norm}
\end{equation}

Equation \ref{equation:trace-norm} implies that the norm of $M(\dot{x})$ serves as a real-valued, totally ordered amplitude of $\dot{x}$. Together with $\| M(\dot{x}) \|_*$, the 
Schatten $2$-norm $\| M(\dot{x}) \|_2$ is also a valid norm of $\dot{x}$. However, due to the non-isometric nature of the Fourier transform, $\| M(\dot{x}) \|_2$ is not equal to the Frobenius norm  $\|\operatorname{tensor}(\dot{x}) \|_{F}$.

The nonnegative t-scalar $|\dot{x}| \doteq \sqrt{\dot{x}^{*} \circ \dot{x}}$ entirely encompasses the amplitude information of the t-scalar $\dot{x}$. Both the Schatten $2$-norm and the tensor's Frobenius norm of $\dot{x}$ can be derived from $|\dot{x}|$ as follows:

The non-negative t-scalar $|\dot{x}| \doteq \sqrt{\dot{x}^{*} \circ \dot{x}}$ fully encompasses the amplitude information of the t-scalar $\dot{x}$. Both the Schatten $2$-norm and the Frobenius norm of the underlying tensor of $\dot{x}$ can be derived from $|\dot{x}|$ as follows:
\begin{equation} 
	\| M(\dot{x}) \|_2 
	= \sqrt{K}\cdot \| \operatorname{tensor}(\dot{x}) \|_F =
	\sqrt{\operatorname{trace} |\dot{x} | }    \;.
	\label{equation:spectral-frobenius-norm-for-tscalars}
\end{equation}
where $K$ is the dimension of the algebra $C$.

\subsection{A Representation Model for T-Matrices and Higher-Order Measures}
Consider a t-matrix $\dot{X} \in C^{D_1\times D_2}$, characterized by its higher-order singular values $\dot{\sigma}_1 \geq \dot{\sigma}_2 \geq \ldots \geq \dot{\sigma}_n \geq \ldots \geq \dot{z}$. The higher order Schatten $p$-norm of $\dot{X}$ is defined by
\begin{equation}
	\mathcal{N}_p(\dot{X}) = 
	\left(
	\sum\limits_{n}  \dot{\sigma}_n^{p}
	\right)^{1/p}  \geq \dot{z} \;.
	\label{equation:higher-order-norm}
\end{equation}

A t-matrix $\dot{X} \in C^{D_1 \times D_2}$ can be represented by placing the diagonal matrix of each corresponding t-scalar into its respective block in the final matrix.
If the diagonal matrix size of a t-scalar is $K\times K$, then the matrix representation of $\dot{X}$ is $\mathbbm{C}^{KD_1\times KD_2}$, as shown by Liao et al.~\cite{liao2022approximation}.

However, matrix representations are not unique. In addition to the above format, a more convenient representation uses the direct sum of matrices, also known as the block diagonal sum. This operation combines several matrices into a larger one, where the summand matrices are arranged along the main diagonal and off-diagonal blocks are filled with zeros.

Given a t-matrix $\dot{X} \in C^{D_1\times D_2}$ with spectral slices denoted by $\tilde{X}_1,\dots,\tilde{X}_K$, the direct sum representation of $\dot{X}$ is established via a bijective mapping:
\begin{equation}
	\dot{X} \mapsto 
	M(\dot{X}) \doteq  \tilde{X}_1 \oplus \tilde{X}_2  \dots \oplus \tilde{X}_K \doteq 
	\begin{bmatrix}
		\tilde{X}_1 & 0 & \cdots & 0 \\
		0 & \tilde{X}_2 & \cdots & 0 \\
		\vdots & \vdots & \ddots & \vdots \\
		0 & 0 & \cdots & \tilde{X}_K   
	\end{bmatrix}
	\;.
\end{equation}

This mapping extends the bijective mapping for t-scalars given in Equation \ref{equationn:matrix-representation-tscallars}. Kilmer et al. proposed the first version of this mapping for the analysis of generalized matrices with order-one entries~\cite{kilmer2011factorization,kilmer2013third}. However, the convenience of its direct sum properties has been largely overlooked by subsequent authors, indicating the underexplored potential of this matrix representation.

It is easy to see that $\dot{X}^{*} \mapsto M(\dot{X})^{H} = \oplus_{k=1}^{K} \tilde{X}_{k}^{H} $. Furthermore, given the one-to-one nature of the mapping, we can define the rank of $\dot{X}$ by the rank of $M(\dot{X})$, which leads to the following equation:
\begin{equation}
	\operatorname{rank} \dot{X} \doteq \operatorname{rank} M(\dot{X}) = \sum\limits_{k=1}^{K}  \operatorname{rank} \tilde{X}_k \;.
\end{equation}

This definition corresponds to the ``trace rank'' discussed in Section \ref{section:Higher-OrderRankandItsTraceVariant}. The direct sum representation also provides a mathematical justification for Algorithm \ref{alg:TensorSVD}. In particular, if the singular value decomposition of the summand $\tilde{X}_{k}$ is $\tilde{X}_{k} = U_{k} \cdot S_{k} \cdot V_{k}^{H}$, the following equation holds for all $\dot{X}$,
\begin{equation}
	\oplus_{k=1}^{K} \tilde{X}_{k} = \oplus_{k=1}^{K} U_{k} \cdot S_{k} \cdot  V_{k}^{H} = 
	\oplus_{k=1}^{K} U_{k} \cdot 
	\oplus_{k=1}^{K} S_{k} \cdot 
	\oplus_{k=1}^{K} V_{k}^{H}  \;.
\end{equation}

In addition, the following equation is valid, which is the TSVD of $\dot{X}$:
\begin{equation}
	\dot{X} = 
	M^{-1}\left(\oplus_{k=1}^{K} U_{k} \right) \circ 
	M^{-1}\left(\oplus_{k=1}^{K} S_{k} \right) \circ 
	M^{-*}\left(\oplus_{k=1}^{K} V_{k} \right)
\end{equation}
where $M^{-*}(\cdot)$ is a shorthand for $(M^{-1}(\cdot))^{*}$.

\subsection{Lagrange Multiplier with t-matrix variables}

The direct sum properties also validate the spectral-slice-wise mechanism exhibited in Algorithms \ref{alg:TensorSVT}, \ref{alg:ADMM}, and \ref{alg:generalized-ADMM}. We are now able to present a generalized version of Equation \ref{equation:lagrange}. The Lagrange multiplier with t-matrix variables has the following form
\begin{equation}
	L(\dot{X}, \dot{E}, \dot{Y}, \tau)= \|\dot{X}\|_{*}+ \langle \dot{Y}, \dot{M} - \dot{X} - \dot{E} \rangle + \frac{1}{2\tau} \| \dot{M} - \dot{X} - \dot{E} \|_2^2  \;.
	\label{equation:genneralized-lagrange}
\end{equation}

The real-valued Schatten $2$-norm $\|\dot{X} \|_2$ can also be derived from the higher-order Schatten $2$-norm $\mathcal{N}_{p}(\dot{X})$ given in Equation \ref{equation:higher-order-norm}. Consequently, for all t-matrices $\dot{X}$, the following equation holds
\begin{equation}
	\|\dot{X} \|_2 \doteq   
	\|M(\dot{X}) \|_2 =  
	\sqrt{  
		\operatorname{trace}  \left( \mathcal{N}_2(\dot{X})^{2} \right)  
	}  \;\geqslant\; 0  \;.
\end{equation}

It is important to note that $\|\dot{X} \|_2$ is measured by spectral slices and the Fourier transform is not isometric. Therefore, unlike the case for conventional matrices, the Schatten $2$-norm $\|\dot{X} \|_2$ and the Frobenius norm $\| \operatorname{tensor} (\dot{X})\|_F$ are different. For any t-matrix $\dot{X}$, the equality 
$\|\dot{X} \|_2 = \sqrt{K} \cdot \| \operatorname{tensor} (\dot{X})\|_F
$ holds.

Similarly, the nuclear norm of any t-matrix $\dot{X}$ is defined by the nuclear norm of $M(\dot{X})$. As with the Schatten $2$-norm norm, the real-valued nuclear norm $\|\dot{X} \|_{*}$ can be derived from its higher-order counterpart $\mathcal{N}_1(\dot{X})$ as follows:
\begin{equation}
	\|\dot{X} \|_{*} \doteq \| M(\dot{X}) \|_{*} = 
	\operatorname{trace}\big( \mathcal{N}_1(\dot{X}) \big) \;.
\end{equation}

Currently, a real inner product $\langle \dot{X}, \dot{Y} \rangle$ of a pair of t-matrices $\dot{X}, \dot{Y}$ is required to have the full Lagrange multiplier as in Equation \ref{equation:genneralized-lagrange}. However, if the inner product $\langle \dot{X}, \dot{Y} \rangle$ is defined as $\operatorname{trace}\big( M(\dot{X})^{H} M(\dot{Y}) \big)$, it will be complex.

To ensure a real-valued inner product instead of a complex-valued one, a feasible strategy is to isomorph $M(\dot{X})$ to a real matrix for any t-matrix $\dot{X}$.

According to representation theory, any complex number $a+ b \sqrt{-1}$ can be represented as a $2\times 2$ real matrix using the following mapping:
\begin{equation} 
	a+ b \sqrt{-1} \mapsto 
	\begin{bmatrix}
		a & -b \\
		b & a \\
	\end{bmatrix}  \;.
\end{equation}

By replacing each complex entry of $M(\dot{X}) \in \mathbbm{C}^{KD_1\times KD_2}$ with its $2\times 2$ real matrix equivalent, the complex matrix $M(\dot{X}) \in \mathbbm{C}^{KD_1\times KD_2}$ can be isomorphically transformed into a real matrix $R(\dot{X}) \in \mathbbm{R}^{2KD_1\times 2KD_2}$. Thus, the real inner product for any pair of t-matrices $\dot{X}$ and $\dot{Y}$ can be defined as
\begin{equation}
	\langle 
	\dot{X},\dot{Y}
	\rangle \doteq (1/2) \cdot \operatorname{trace} \big(R(\dot{X})^{T}  R(\dot{Y}) \big)  \;.
\end{equation}
The coefficient $(1/2)$ is essential to account for the doubling of absolute values in the real representation.

The following discussion is standard in convex analysis with the Lagarange multiplier given by Equation \ref{equation:genneralized-lagrange}  and has been studied extensively by previous authors. Therefore, it is beyond the scope of this appendix.

\end{document}